# Agentic retrieval-augmented reasoning reshapes collective reliability under model variability in radiology question answering

Mina Farajiamiri[†] (1), Jeta Sopa[†] (2), Saba Afza (2), Lisa Adams (3), Felix Barajas Ordonez (1,4), Tri-Thien Nguyen (2,5), Mahshad Lotfinia (1,4), Sebastian Wind (2,6), Keno Bressem (3,7), Sven Nebelung (1,4), Daniel Truhn (1,4), Soroosh Tayebi Arasteh (1,4,8,9)

(1) Lab for AI in Medicine, RWTH Aachen University, Aachen, Germany.
(2) Pattern Recognition Lab, Friedrich-Alexander-Universität Erlangen-Nürnberg, Erlangen, Germany.
(3) Department of Diagnostic and Interventional Radiology, TUM University Klinikum, School of Medicine and Health, Klinikum rechts der Isar, Technical University of Munich, Munich, Germany.
(4) Department of Diagnostic and Interventional Radiology, University Hospital RWTH Aachen, Aachen, Germany.
(5) Institute of Radiology, University Hospital Erlangen, Erlangen, Germany.
(6) Erlangen National High Performance Computing Center, Friedrich-Alexander-Universität Erlangen-Nürnberg, Erlangen, Germany.
(7) Department of Cardiovascular Radiology and Nuclear Medicine, TUM University Clinic, School of Medicine and Health, German Heart Center, Technical University of Munich, Munich, Germany.
(8) Department of Urology, Stanford University, Stanford, CA, USA.
(9) Department of Radiology, Stanford University, Stanford, CA, USA.

[†] Mina Farajiamiri and Jeta Sopa are shared-first-authors.

**Correspondence**
Soroosh Tayebi Arasteh, Dr.-Ing., Dr. rer. medic.
Lab for AI in Medicine
Department of Diagnostic and Interventional Radiology
University Hospital RWTH Aachen
Pauwelsstr. 30, 52074 Aachen, Germany
Email: soroosh.arasteh@rwth-aachen.de

**Abstract**

Agentic retrieval-augmented reasoning pipelines are increasingly used to structure how large language models (LLMs) incorporate external evidence in clinical decision support. These systems iteratively retrieve curated domain knowledge and synthesize it into structured reports before answer selection. Although such pipelines can improve performance, their impact on reliability under model variability remains unclear. In real-world deployment, heterogeneous models may align, diverge, or synchronize errors in ways not captured by average accuracy. We evaluated 34 LLMs on 169 expert-curated publicly available radiology questions, comparing zero-shot inference with a radiology-specific multi-step agentic retrieval condition in which all models received identical structured evidence reports derived from curated radiology knowledge. Agentic inference reduced inter-model decision dispersion (median entropy 0.48→0.13; $P=5.6 \times 10^{-9}$) and increased robustness of correctness across models (mean 0.74→0.81; $P=5.6 \times 10^{-9}$). Majority consensus also increased overall ($P=2.9 \times 10^{-5}$). Consensus strength and robust correctness remained strongly correlated under both inference strategies ($\rho=0.88$ for zero-shot; $\rho=0.87$ for agentic), although high agreement did not guarantee correctness. Rare high-consensus, low-robustness failures occurred under both methods (1% vs. 2%). Response verbosity showed no meaningful association with correctness. Among 572 incorrect outputs, 72% were associated with moderate or high clinically assessed severity, although inter-rater agreement was low ($\kappa=0.02$). Agentic retrieval therefore was associated with more concentrated decision distributions, stronger consensus, and higher cross-model robustness of correctness. However, coordinated failures and clinically consequential error modes persisted. These findings suggest that evaluating agentic systems through accuracy or agreement alone may not always be sufficient, and that complementary analyses of stability, cross-model robustness, and potential clinical impact are needed to characterize reliability under model variability.

# 1. Introduction

Large language models (LLMs) are increasingly incorporated into decision-making pipelines across science, engineering, and healthcare, where their outputs can shape expert reasoning, downstream actions, and risk-bearing outcomes[1–7]. In clinical domains such as radiology[1], recent progress in retrieval-augmented and multi-step reasoning systems has shown improvements in performance on knowledge-intensive tasks[8]. However, improvements in mean accuracy alone are insufficient to characterize reliability in real deployments, where systems vary across architectures, vendors, versions, and operational constraints[9,10]. A central but underexplored question is therefore not only whether an LLM-based system is correct on average, but whether its decisions are stable and reproducible when the deployed model changes. This framing motivates a focus on model variability as a first-class reliability dimension. In practice, the model is rarely fixed: organizations may switch providers, roll out new versions, or route queries across different backends to meet latency and cost constraints. Under such variability, a decision pipeline can appear reliable in aggregate while remaining fragile if answers depend strongly on model choice. From a safety perspective, inter-model variability is not merely noise to be averaged away; it can reveal instability, sensitivity to context, and failure models that are masked by reporting only mean performance[11,12].

Agentic reasoning introduces competing forces that make stability difficult to predict a priori. Shared retrieval sources and structured templates may align models toward similar conclusions, reducing dispersion and increasing apparent agreement[13,14]. Conversely, the same structure can synchronize errors: if retrieved evidence is misleading or intermediate reasoning channels attention toward the wrong features, multiple models may converge on the same incorrect answer[15]. Such coordinated failures are concerning in high-stakes settings because they can produce false confidence through apparent consensus[16]. It remains unclear how agentic retrieval and reasoning affect inter-model agreement, whether shifts in agreement track correctness, whether correctness becomes more robust across heterogeneous models, and whether consensus remains a reliable indicator of validity once models are exposed to shared structured evidence[11,17,18]. In parallel, confidence signaling remains an unresolved challenge[11,17]. Users are frequently exposed to proxies such as explanation length, reasoning verbosity, or structured rationales. Yet it is unclear whether these signals reliably correlate with correctness or safety under agentic inference. If agentic systems produce longer or more detailed outputs without improving the alignment between such proxies and correctness, they may increase over-trust rather than reliability[19,20]. Moreover, even when collective behavior appears more coordinated or more robust, the clinical severity of residual errors may remain heterogeneous and safety-relevant[10,21,22].

To address these gaps, we present a controlled evaluation framework for cross-model reliability under a shared-evidence setting, using a standardized agentic retrieval-augmented[8,13,23] pipeline to hold retrieval and evidence synthesis constant (**Figure 1**). Rather than treating accuracy as a single endpoint, we decompose reliability into complementary dimensions[10,11]: inter-model decision stability, majority-consensus behavior, robustness of correctness across models, coupling between agreement and correctness, relationships between verbosity-based confidence proxies and decision validity, and clinically assessed severity of incorrect answer options[15]. We treat these dimensions as distinct but interacting axes of system behavior, allowing us to distinguish concentration of decisions from correctness, robustness from consensus, and frequency of error from potential clinical impact[18,24]. We apply this framework to a heterogeneous panel of 34 LLMs spanning proprietary and open-weight systems, parameter scales from small to very large, and both general-purpose and medically adapted models. The panel includes models such as those from the OpenAI, Qwen, Llama, DeepSeek, Gemma, Claude, Gemini, and Mistral families, reflecting realistic deployment diversity rather than a single vendor or architecture. These models are evaluated on 169 expert-curated radiology questions from two datasets: the Benchmark radiology question answering dataset (Benchmark-RadQA; n = 104), sourced from the RadioRAG study[23], and a board-style radiology question answering dataset (Board-RadQA; n = 65), sourced from the RaR study[8]. For each question, we compare zero-shot inference with a standardized agentic retrieval-augmented condition[8]. In the zero-shot condition, the model receives only the question stem and answer options. In the agentic condition, the model receives additional structured evidence report produced by an orchestration pipeline[25]. The pipeline retrieves clinically relevant information from a curated radiology knowledge base[26] and synthesizes it into an informative yet neutral report about the question and the corresponding options. The orchestration process is held fixed across all models, and for a given question all models receive identical retrieved context, allowing us to isolate how different models behave when exposed to the same structured evidence rather than conflating model differences with retrieval or planning differences.

Our analyses ask: Does agentic inference reduce inter-model decision entropy, indicating increased concentration of decisions across models[18,27]? When agreement changes, does it preferentially amplify correct or incorrect majorities? Does agentic inference increase robustness of correctness, defined as the fraction of models that independently reach the correct answer[28]? How tightly are consensus strength and correctness coupled under zero-shot vs. agentic conditions, and can high agreement coexist with fragile or incorrect outcomes[11,16]? Do verbosity-based confidence proxies meaningfully track correctness under either inference strategy[19,20]? Finally, what is the distribution and inter-rater reliability[29] of clinically assessed error severity, and how does this safety-relevant dimension relate to collective decision structure[10]? By reframing evaluation around stability, robustness, coupling, and clinical impact, this work advances a structured and safety-aware assessment of LLM-based decision support systems in radiology and other high-stakes domains.

## 2. Results

We evaluated zero-shot inference and agentic retrieval-augmented reasoning across 169 multiple-choice radiology questions drawn from the Benchmark-RadQA (n = 104) and Board-RadQA (n = 65) datasets (**Supplementary Table S1**), answered by 34 LLMs (model specifications in **Supplementary Table S2**). The primary data comprised per-question discrete answer choices from each model under each inference condition, enabling paired, question-level comparisons. Pooled paired analyses across all 169 questions constitute the primary confirmatory comparisons, while dataset-stratified results are reported, in **Supplementary Note 1**, to assess directional consistency across subsets. In accordance with the prespecified outcome hierarchy, inter-model decision stability and robustness of correctness represent the primary endpoints, while consensus behavior, coupling metrics, verbosity analyses, and severity annotations are interpreted as secondary or exploratory characterizations of collective behavior. For descriptive context, **Table 1** reports single-model accuracy under zero-shot and agentic inference for each of the 34 LLMs.

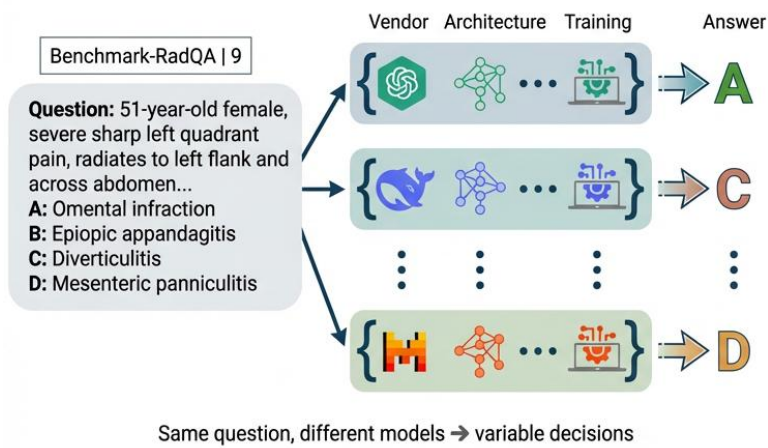
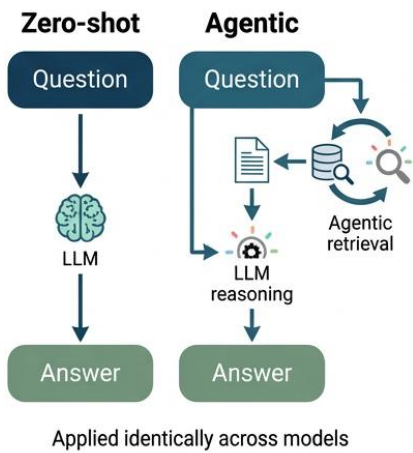
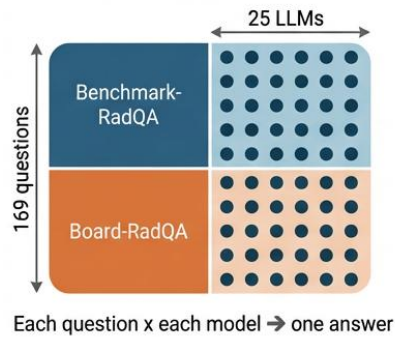
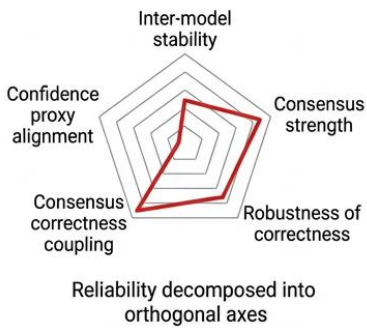
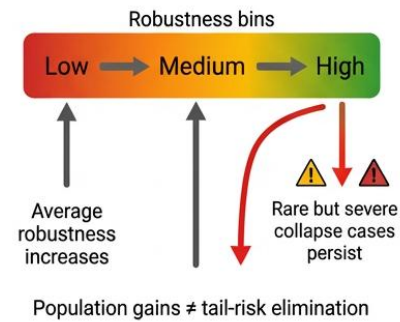

**Figure 1:** Study overview and experimental design. **a** A single radiology multiple-choice question is presented to a heterogeneous panel of large language models (LLMs) that differ in vendor, architecture, and training. Although all models receive the same input, their final answers may diverge, illustrating model variability and motivating evaluation of collective reliability rather than single-model accuracy. **b** Zero-shot inference consists of direct question-to-answer generation by each model. In contrast, the agentic retrieval-augmented pipeline incorporates structured multi-step reasoning and iterative evidence retrieval before answer selection. Both pipelines are applied identically across all 34 models to isolate the effect of inference strategy under controlled conditions. **c** A total of 169 multiple-choice radiology questions (Benchmark-RadQA, n = 104; Board-RadQA, n = 65) are answered independently by 34 LLMs under both zero-shot and agentic conditions. Each question-model pair yields one discrete answer, enabling paired per-question comparisons across inference strategies. **d** Collective behavior is decomposed into orthogonal metrics: inter-model stability (entropy), consensus strength (majority fraction), robustness of correctness (fraction of models correct), coupling between consensus and correctness, and alignment between verbosity and correctness as a confidence proxy. This multidimensional framework separates coordination structure from validity. **e** Robustness scores are stratified into low, medium, and high bins to characterize reliability regimes. Agentic reasoning is evaluated for both population-level robustness gains and the persistence of rare but severe collapse cases, emphasizing that improvements in average robustness do not necessarily eliminate tail-risk failure modes.

**Table 1:** Accuracy of language models across zero-shot prompting and the agentic model over all 169 questions. Accuracy is reported in percentage as mean ± standard deviation, with 95% confidence intervals shown in brackets. Results are based on 169 questions, using bootstrapping with 1,000 repetitions and replacement while preserving pairing. P-values were calculated for each model using McNemar's test on paired outcomes between the zero-shot and agentic methods, and adjusted for multiple comparisons using the false discovery rate. A p-value < 0.05 was considered statistically significant. Accuracy is presented alongside total correct answers per method.

| Model name | Zero-shot | | Agentic | | P-value |
| --- | --- | --- | --- | --- | --- |
| | Accuracy (%) | Total correct (n) | Accuracy (%) | Total correct (n) | |
| Claude-Sonnet-4.6 | 85 ± 3 [80, 90] | 144 | 88 ± 3 [82, 92] | 148 | 0.577 |
| Gemini-3.1-Pro | 82 ± 3 [76, 87] | 138 | 91 ± 2 [86, 95] | 154 | 0.012 |
| MiniMax-M2.5 | 79 ± 3 [73, 86] | 134 | 80 ± 3 [74, 86] | 135 | > 0.999 |
| GLM-5 | 82 ± 3 [76, 88] | 139 | 85 ± 3 [80, 91] | 144 | 0.527 |
| LFM2.5-1.2B-Thinking | 56 ± 4 [49, 63] | 94 | 76 ± 3 [69, 82] | 129 | < 0.001 |
| Kimi-K2.5 | 83 ± 3 [78, 89] | 141 | 85 ± 3 [79, 91] | 144 | 0.729 |
| Palmyra-X5 | 86 ± 3 [81, 91] | 146 | 86 ± 3 [80, 91] | 145 | 1.000 |
| MiMo-V2-Flash | 79 ± 3 [72, 85] | 133 | 87 ± 3 [82, 92] | 147 | 0.063 |
| Llama4-Scout-16E | 82 ± 3 [76, 88] | 139 | 85 ± 3 [79, 90] | 143 | 0.585 |
| Llama3.3-8B | 69 ± 4 [63, 76] | 117 | 75 ± 3 [67, 80] | 126 | 0.275 |
| Llama3.3-70B | 80 ± 3 [74, 86] | 135 | 86 ± 3 [80, 91] | 145 | 0.144 |
| Llama3-Med42-8B | 63 ± 4 [56, 70] | 107 | 74 ± 4 [67, 80] | 125 | 0.024 |
| Llama3-Med42-70B | 75 ± 3 [69, 82] | 127 | 79 ± 3 [73, 85] | 134 | 0.417 |
| DeepSeek R1-70B | 84 ± 3 [79, 89] | 142 | 84 ± 3 [78, 89] | 142 | 1.000 |
| DeepSeek-R1 | 88 ± 2 [82, 92] | 148 | 84 ± 3 [78, 89] | 142 | 0.343 |
| DeepSeek-V3 | 84 ± 3 [79, 89] | 142 | 89 ± 3 [83, 93] | 150 | 0.218 |
| GPT-5 | 84 ± 3 [78, 89] | 142 | 87 ± 3 [82, 92] | 147 | 0.402 |
| GPT-5.2 | 79 ± 3 [73, 85] | 134 | 87 ± 3 [82, 92] | 147 | 0.035 |
| o3 | 85 ± 3 [79, 90] | 144 | 89 ± 2 [84, 93] | 150 | 0.239 |
| GPT-3.5-turbo | 64 ± 4 [57, 71] | 108 | 77 ± 3 [70, 83] | 130 | 0.012 |
| GPT-4-turbo | 77 ± 3 [70, 83] | 130 | 82 ± 3 [76, 88] | 138 | 0.277 |
| Mistral Large (123B) | 81 ± 3 [75, 87] | 137 | 86 ± 3 [80, 91] | 146 | 0.166 |
| Ministral-8B | 54 ± 4 [46, 62] | 91 | 76 ± 3 [69, 82] | 128 | < 0.001 |
| MedGemma-4B-it | 64 ± 4 [57, 71] | 108 | 75 ± 3 [68, 81] | 127 | 0.024 |
| MedGemma-27B-text-it | 79 ± 3 [72, 85] | 133 | 84 ± 3 [78, 89] | 142 | 0.220 |
| Gemma-3-4B-it | 51 ± 4 [44, 59] | 86 | 71 ± 3 [64, 78] | 121 | < 0.001 |
| Gemma-3-27B-it | 71 ± 3 [64, 78] | 120 | 83 ± 3 [78, 89] | 141 | 0.009 |
| Qwen3-8B | 75 ± 3 [68, 81] | 127 | 81 ± 3 [75, 86] | 137 | 0.218 |
| Qwen3-235B | 86 ± 3 [81, 91] | 146 | 85 ± 3 [79, 91] | 144 | 0.877 |
| Qwen2.5-0.5B | 37 ± 4 [30, 45] | 63 | 48 ± 4 [41, 56] | 81 | 0.079 |
| Qwen2.5-3B | 63 ± 4 [57, 71] | 107 | 73 ± 3 [66, 80] | 124 | 0.038 |
| Qwen2.5-7B | 63 ± 4 [57, 70] | 107 | 78 ± 3 [71, 84] | 132 | 0.002 |
| Qwen2.5-14B | 74 ± 3 [68, 81] | 126 | 79 ± 3 [73, 85] | 134 | 0.354 |
| Qwen2.5-70B | 79 ± 3 [73, 85] | 134 | 84 ± 3 [78, 89] | 142 | 0.229 |

## 2.1. Agentic reasoning alters inter-model decision stability

We first quantified how agentic retrieval-augmented reasoning[8] changes inter-model decision stability, operationalized as the Shannon entropy[27] of the answer distribution across the model panel for each question. Across all 169 questions, agentic reasoning yielded lower entropy than zero-shot inference (**Table 2, Figure 2**), indicating that model decisions were more concentrated under the agentic pipeline. The median entropy decreased from 0.48 (IQR 0.55) under zero-shot to 0.13 (IQR 0.51) under agentic reasoning, and the mean decreased from 0.50 to 0.40. The paired per-question shift was significant ($P = 5.6 \times 10^{-9}$); rank-biserial r = −0.93), with a median ΔH of −0.13 and mean ΔH of −0.19, demonstrating an overall reduction in dispersion across models. Per-question comparisons showed that the effect was not uniform. In the pooled dataset, entropy decreased in 115 of 169 questions (68%), increased in 31 (18%), and remained unchanged in 23 (14%). **Supplementary Table S3** provides representative per-question examples illustrating both stabilizing cases (negative ΔH) and occasional destabilizations (positive ΔH). Importantly, entropy captures coordination rather than correctness: lower entropy reflects stronger alignment among models, not necessarily higher validity. This distinction motivates subsequent analyses examining whether increased stability translates into more robust correctness or more reliable consensus behavior under model variability.

## 2.2. Changes in consensus strength do not reliably track correctness

We next tested whether agentic retrieval-augmented reasoning changes the strength of inter-model consensus, and whether shifts in consensus preferentially align with correct decisions. For each question and method, we computed the majority fraction, defined as the proportion of models selecting the modal answer option, and evaluated whether the resulting majority decision matched the reference standard (**Table 3, Supplementary Table S4, Figure 3**). Across all 169 questions, agentic reasoning increased consensus strength. The median majority fraction increased from 0.85 (IQR 0.21) under zero-shot inference to 0.97 (IQR 0.18) under agentic reasoning. The paired per-question comparison, restricted to non-zero ΔM pairs, showed a significant shift ($P = 2.9 \times 10^{-5}$), with a positive median ΔM of 0.03, indicating that agreement typically increased under the agentic pipeline. Per-question categorization showed that agreement increased with a correct majority in 95 of 169 questions (56%), increased with an incorrect majority in 11 (7%), decreased in 30 (18%), and remained unchanged in 33 (20%). Thus, agreement amplification occurred more often than agreement loss and was more frequently associated with correct majorities, but it was not exclusively correctness-favorable. In a measurable subset of cases, agentic reasoning concentrated models around an incorrect answer. These findings indicate that agentic retrieval strengthens inter-model consensus at the population level, yet consensus strength remains an imperfect indicator of decision validity.

**Table 2:** Inter-model decision stability under zero-shot vs. agentic inference. Inter-model decision stability was quantified for each question using Shannon entropy of the answer distribution across the 34-model panel, with lower entropy indicating stronger agreement among models. Entropy values are reported for zero-shot and agentic retrieval-augmented inference across pooled questions and dataset-specific subsets. Paired changes in decision stability (ΔH) were computed on a per-question basis as agentic minus zero-shot entropy; negative ΔH values indicate increased stability under agentic inference, whereas positive values indicate decreased stability. Summary statistics are reported as medians with interquartile ranges and as means. The distribution of per-question entropy changes is additionally summarized by the number and proportion of questions exhibiting decreased (ΔH < 0), increased (ΔH > 0), or unchanged (ΔH = 0) entropy. Statistical significance of paired entropy changes was assessed using a two-sided Wilcoxon signed-rank test on paired per-question entropy values, with rank-biserial correlation (r) reported as an effect size for the paired comparison.

| Metric | Pooled datasets | Board-RadQA dataset | Benchmark-RadQA dataset |
|---|---|---|---|
| Questions [n] | 169 | 65 | 104 |
| Entropy (zero-shot), median (IQR) | 0.48 (0.55) | 0.35 (0.50) | 0.53 (0.56) |
| Entropy (agentic), median (IQR) | 0.13 (0.51) | 0.13 (0.13) | 0.24 (0.57) |
| Entropy (zero-shot), mean | 0.50 | 0.41 | 0.55 |
| Entropy (agentic), mean | 0.40 | 0.20 | 0.38 |
| Paired entropy change (ΔH), median | −0.13 | −0.13 | −0.13 |
| Paired entropy change (ΔH), mean | −0.19 | −0.20 | −0.18 |
| Questions with ΔH < 0 / > 0 / = 0, n (%) | 115 (68%) / 31 (18%) / 23 (14%) | 46 (71%) / 8 (12%) / 11 (17%) | 69 (66%) / 23 (22%) / 12 (12%) |
| P-value | $5.6 \times 10^{-9}$ | $9.6 \times 10^{-6}$ | $9.4 \times 10^{-9}$ |
| Rank-biserial correlation (r) | -0.93 | -0.94 | -0.64 |

## 2.3. Agentic reasoning increases robustness of correctness across models

We next evaluated whether agentic retrieval-augmented reasoning improves the robustness of correctness across model variability. Robustness was defined, for each question, as the fraction of models producing the correct answer, capturing sensitivity to model choice rather than average performance (**Figure 4, Supplementary Table S5**). Across all 169 questions, agentic reasoning produced a clear upward shift in robustness. Mean robustness increased from 0.74 under zero-shot inference to 0.81 under agentic reasoning, and the median increased from 0.79 to 0.94. The proportion of questions in the high-robustness bin rose from 50% to 72%, while the medium-robustness fraction declined from 41% to 17% and the low-robustness fraction increased from 9% to 11%. The paired per-question change was statistically significant (P = $5.6 \times 10^{-9}$; rank-biserial r = 0.45), with a median ΔR of +0.07, corresponding to approximately one additional model answering correctly per question (**Supplementary Table S6**). These results indicate that agentic reasoning increases cross-model reproducibility of correct decisions at the population level.

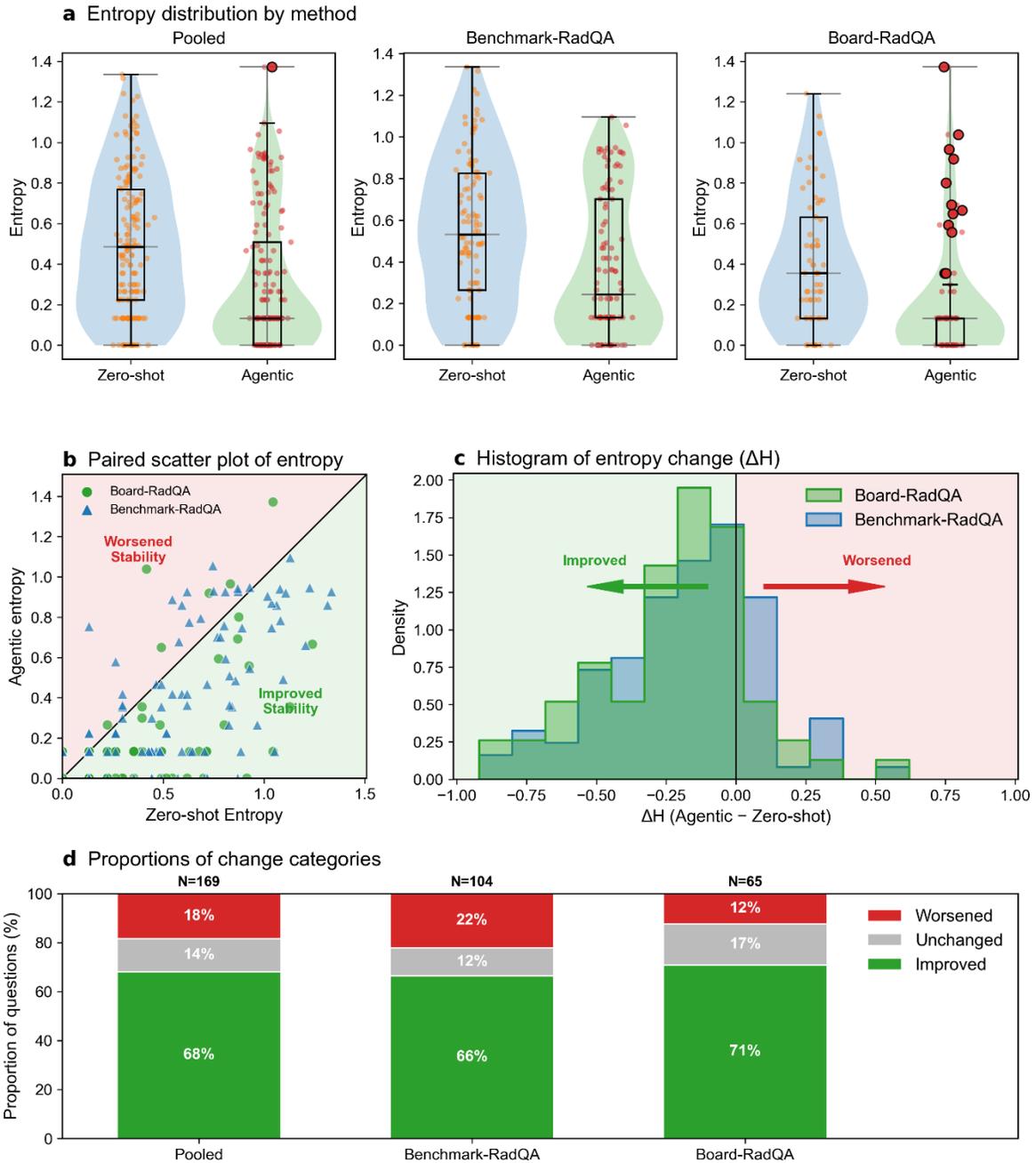

**Figure 2:** Inter-model decision stability under zero-shot vs. agentic retrieval-augmented inference. Inter-model stability was quantified for each question using Shannon entropy of the answer distribution across the 34-model panel, with lower entropy indicating stronger concentration of decisions. Results are shown for pooled questions (n = 169) and separately for Benchmark-RadQA (n = 104) and Board-RadQA (n = 65). **a** Violin plots with embedded boxplots display entropy distributions under zero-shot and agentic inference for pooled and dataset-specific subsets. Points represent per-question entropy values; boxplots show median and interquartile range. **b** Paired scatter plot of per-question entropy values under zero-shot (x-axis) vs. agentic inference (y-axis). The identity line indicates no change; points below the line reflect reduced entropy (improved stability) under agentic inference, whereas points above indicate increased entropy. **c** Histograms of per-question entropy change (ΔH = agentic − zero-shot) for each dataset. Negative ΔH values correspond to increased stability, and positive values indicate decreased stability. **d** Proportions of questions exhibiting improved (ΔH < 0), unchanged (ΔH = 0), or worsened (ΔH > 0) stability in pooled and dataset-specific analyses. Statistical significance of paired entropy changes was assessed using two-sided Wilcoxon signed-rank tests on per-question entropy values.

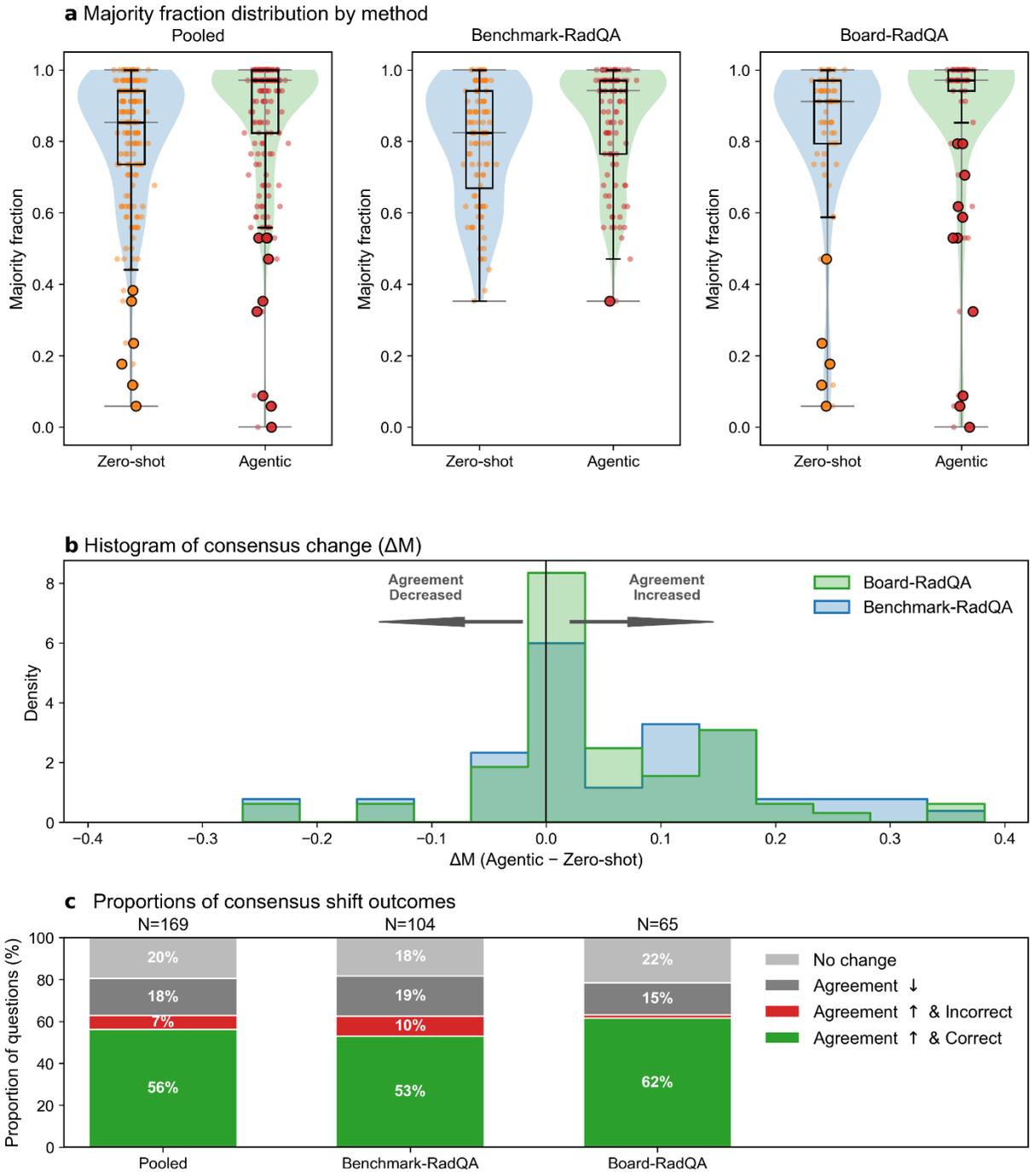

**Figure 3:** Inter-model consensus strength and correctness under zero-shot vs. agentic inference. Inter-model agreement was quantified for each question as the majority fraction (M), defined as the proportion of the 34 models selecting the modal answer option. Results are shown for pooled questions (n = 169) and separately for Benchmark-RadQA (n = 104) and Board-RadQA (n = 65). **a** Violin plots with embedded boxplots display the distribution of majority fractions under zero-shot and agentic retrieval-augmented inference. Points represent per-question values; boxplots indicate median and interquartile range. **b** Histograms of paired changes in agreement strength (ΔM = agentic − zero-shot). Positive ΔM values indicate increased consensus under agentic inference, whereas negative values indicate decreased consensus. Distributions are shown separately for each dataset. **c** Proportions of questions categorized by consensus shift outcome: agreement increased with correct majority, agreement increased with incorrect majority, agreement decreased, or no change. Statistical significance of paired agreement changes was assessed using a two-sided Wilcoxon signed-rank test on non-zero ΔM pairs.

## 2.4. Robustness gains coexist with rare but severe collapse cases

Although robustness most often improved or remained stable under agentic reasoning, a small subset of questions exhibited pronounced decreases (**Supplementary Table S5**). Across all 169 questions, robustness improved in 45 (27%), remained unchanged in 111 (66%), and decreased in 10 (7%) (**Supplementary Table S6**). Some decreases were large in magnitude, including collapse events with ΔR = −0.79, reflecting coordinated shifts in correctness across many models rather than isolated failures. While upward transitions dominated overall, these tail events demonstrate that agentic reasoning can synchronize incorrect decisions in rare cases. Thus, agentic inference improves robustness substantially more often than it degrades it, yet rare robustness collapses persist and represent safety-relevant failure modes.

## 2.5. Output verbosity is a weak and inconsistent proxy for correctness

We next examined whether response length relates to correctness within each inference condition and dataset (**Supplementary Table S7, Supplementary Figure S1**). All comparisons were performed within inference method to isolate verbosity-correctness associations from systematic length differences introduced by the agentic pipeline. Across all 169 questions, verbosity showed only a minimal association with correctness under zero-shot inference and no meaningful association under agentic inference. Under zero-shot inference, the median verbosity was slightly higher for correct responses than for incorrect responses, 280 vs. 256 tokens, with a significant but negligible effect size (P = 0.020; Cliff's δ = 0.04). Although significant, the effect size was negligible, indicating only a very small difference in response length between correct and incorrect outputs. Under agentic inference, median verbosity was nearly identical between correct and incorrect responses 660 vs. 668 tokens (P = 0.833; Cliff's δ = −0.004), with a negligible effect size and no evidence of a relationship between response length and correctness.

## 2.6. Consensus strength and robust correctness are only partially coupled

We examined whether stronger inter-model consensus corresponds to more robust correctness across models by relating majority fraction to robustness at the per-question level (**Table 4, Figure 5**). Across all 169 questions, consensus strength and robustness were coupled under both zero-shot inference ($\rho = 0.88$, $P = 1.8 \times 10^{-55}$) and agentic reasoning ($\rho = 0.87$, $P = 6.8 \times 10^{-54}$).

At the majority-decision level, correct majorities exhibited higher agreement than incorrect majorities. In the pooled zero-shot analysis, the median majority fraction was 0.88 for correct vs. 0.56 for incorrect majorities ($P = 6.1 \times 10^{-8}$; Cliff's δ = 0.82, large). Under agentic inference, the corresponding medians were 0.97 vs. 0.59 ($P = 1.0 \times 10^{-10}$; Cliff's δ = 0.87, large). However, high agreement did not guarantee correctness. We identified four cases of high-consensus, low-robustness behavior: one under zero-shot (1/169, 1%) and three under agentic inference (3/169, 2%) (**Supplementary Table S8**). Thus, consensus and robustness are strongly aligned on average under both strategies, yet coordinated incorrect convergence can still occur.

**Table 3:** Majority-vote agreement strength and correctness under zero-shot vs. agentic reasoning. Inter-model agreement strength was quantified for each question as the majority fraction, defined as the proportion of models selecting the modal answer option among the 34-model panel. Values are reported for zero-shot and agentic retrieval-augmented inference across pooled questions and dataset-specific subsets. Paired changes in agreement strength (ΔM) were computed on a per-question basis as agentic minus zero-shot majority fraction. Positive ΔM indicates increased agreement under agentic reasoning, while negative ΔM indicates decreased agreement. Questions were categorized according to whether agreement increased with a correct majority, increased with an incorrect majority, decreased, or remained unchanged. Statistical significance of paired agreement changes was assessed using a two-sided Wilcoxon signed-rank test applied to non-zero ΔM pairs only. All reported p-values correspond to this paired non-parametric test.

| Metric | Pooled datasets | Board-RadQA dataset | Benchmark-RadQA dataset |
|---|---|---|---|
| Questions [n] | 169 | 65 | 104 |
| Median majority fraction (zero-shot) | 0.85 | 0.91 | 0.82 |
| Median majority fraction (agentic) | 0.97 | 0.97 | 0.94 |
| Median ΔM (agentic − zero-shot) | 0.03 | 0.03 | 0.06 |
| IQR majority fraction (zero-shot) | 0.21 | 0.18 | 0.27 |
| IQR majority fraction (agentic) | 0.18 | 0.06 | 0.21 |
| Questions with agreement ↑ and majority correct | 95 (56%) | 40 (62%) | 55 (53%) |
| Questions with agreement ↑ and majority incorrect | 11 (7%) | 1 (2%) | 10 (10%) |
| Questions with agreement ↓ | 30 (18%) | 10 (15%) | 20 (19%) |
| Questions with no change in agreement | 33 (20%) | 14 (22%) | 19 (18%) |
| Non-zero ΔM pairs | 136 | 51 | 85 |
| P-value | $2.9 \times 10^{-5}$ | $1.5 \times 10^{-4}$ | $5.7 \times 10^{-7}$ |

## 2.7. Clinical severity of incorrect decisions reveals heterogeneous and safety-relevant error patterns

To characterize the clinical risk profile of model failures, we analyzed radiologist-assigned severity labels for all incorrect answer options across all 169 questions (**Figure 6**). Severity annotation was performed independently by two board-certified radiologists (L.A. and T.T.N.) and one senior radiology resident (F.B.O.), all blinded to model identities, inference strategies, and collective performance metrics. For each incorrect answer option, raters classified the anticipated clinical consequence as low, moderate, or high severity (**Supplementary Table S9**).

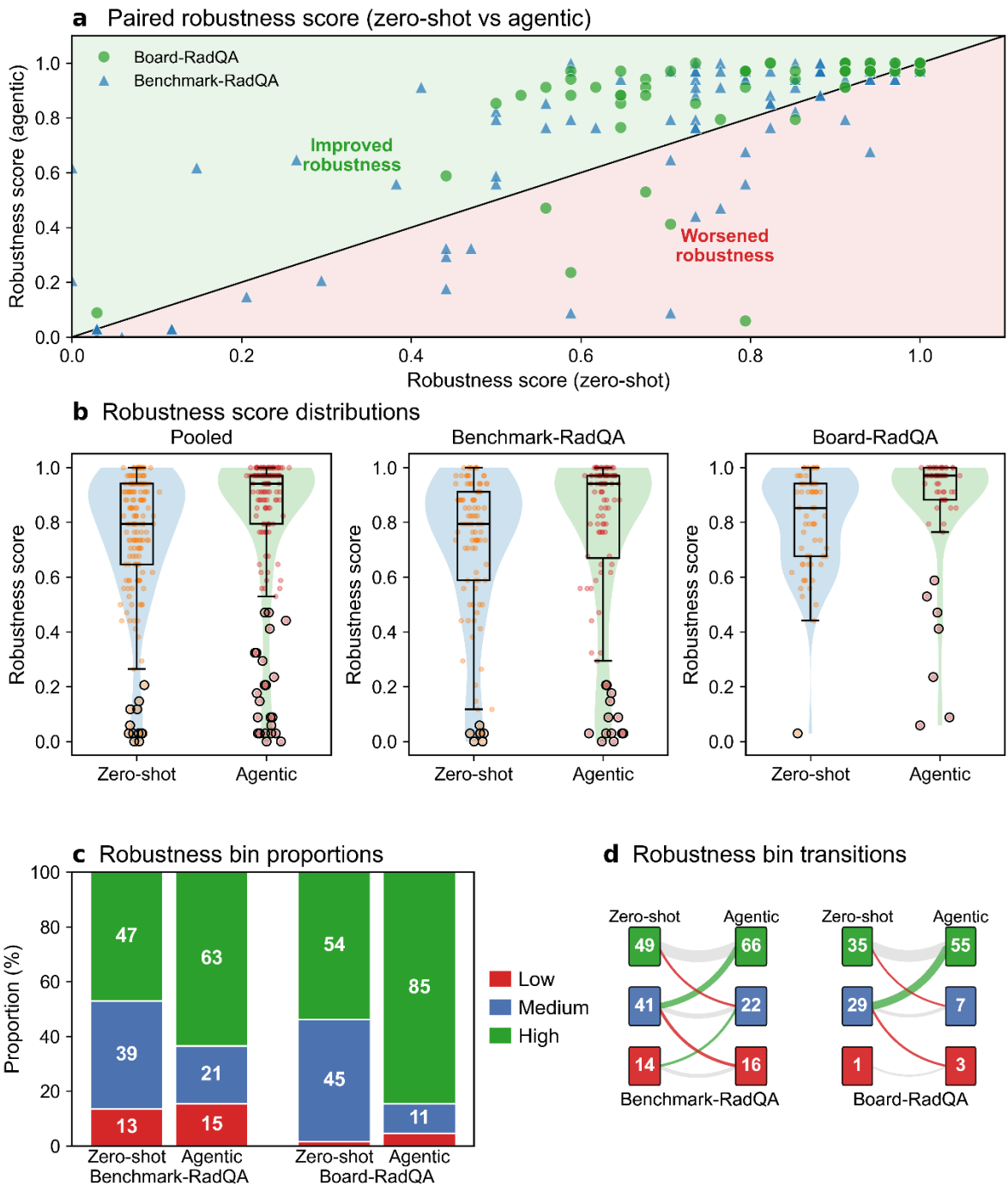

**Figure 4:** Robustness of correctness across models under zero-shot and agentic inference. Robustness (R) was defined per question as the fraction of the 34-model panel selecting the reference-standard answer. **a** Per-question robustness under zero-shot (x-axis) vs. agentic inference (y-axis), shown separately for Benchmark-RadQA and Board-RadQA. The identity line indicates no change. Points above the line reflect improved robustness under agentic inference, whereas points below indicate decreased robustness. **b** Violin plots with embedded boxplots showing the distribution of robustness scores under zero-shot and agentic inference for pooled questions and for each dataset separately. Points represent per-question values; boxplots indicate median and interquartile range. **c** Proportion of questions falling into low, medium, and high robustness bins under zero-shot and agentic inference for each dataset. **d** Transition diagrams illustrating per-question shifts between robustness categories from zero-shot to agentic inference in Benchmark-RadQA and Board-RadQA. Flows depict movements between bins, highlighting dominant medium-to-high transitions as well as rare downward shifts.

**Table 4:** Relationship between inter-model consensus strength and correctness under zero-shot and agentic reasoning. This table summarizes how inter-model consensus strength relates to correctness and robust correctness across zero-shot and agentic inference, reported for pooled questions and dataset-specific subsets (Board-RadQA and Benchmark-RadQA). In the upper section, consensus strength is quantified for each question as the majority fraction (M), defined as the proportion of models selecting the modal answer option among the 34-model panel. Questions are stratified according to whether the majority decision matches the reference-standard answer ("Correct") or not ("Incorrect"). Median and mean majority fractions are reported for each stratum. Statistical comparisons between correct-majority and incorrect-majority questions are performed within each method and dataset using a two-sided Mann-Whitney U test, with Cliff's delta (δ) reported as an effect size. All p-values in the upper section correspond to these unpaired rank-based comparisons. In the lower section, the coupling between consensus strength and robust correctness is quantified using Spearman's rank correlation coefficient (ρ), computed across questions for each inference method and dataset. Robust correctness is defined as the fraction of models selecting the ground-truth answer. Statistical significance of the monotonic association is assessed using two-sided Spearman correlation tests. All p-values in the lower section correspond to tests of the null hypothesis $\rho = 0$.

| Metric | Pooled dataset | | Board-RadQA dataset | | Benchmark-RadQA dataset | |
|---|---|---|---|---|---|---|
| | Zero-shot | Agentic | Zero-shot | Agentic | Zero-shot | Agentic |
| **Majority agreement strength stratified by correctness** | | | | | | |
| Correct [n] | 153 | 149 | 61 | 59 | 92 | 90 |
| Incorrect [n] | 16 | 20 | 4 | 6 | 12 | 14 |
| Median M (Correct) | 0.88 | 0.97 | 0.91 | 0.97 | 0.85 | 0.96 |
| Median M (Incorrect) | 0.56 | 0.59 | 0.15 | 0.21 | 0.58 | 0.67 |
| Mean M (Correct) | 0.85 | 0.91 | 0.88 | 0.95 | 0.82 | 0.89 |
| Mean M (Incorrect) | 0.48 | 0.53 | 0.15 | 0.27 | 0.50 | 0.68 |
| P-value (Mann-Whitney U) | $6.1 \times 10^{-8}$ | $1.0 \times 10^{-10}$ | $8.4 \times 10^{-4}$ | $2.8 \times 10^{-5}$ | $6.8 \times 10^{-5}$ | $3.3 \times 10^{-6}$ |
| Cliff's δ | 0.82 | 0.87 | 1.00 | 0.99 | 0.71 | 0.77 |
| Effect size | Large | Large | Large | Large | Large | Large |
| **Rank-based association between consensus strength and robustness** | | | | | | |
| Questions [n] | 169 | 169 | 65 | 65 | 104 | 104 |
| Spearman ρ | 0.88 | 0.87 | 0.81 | 0.69 | 0.93 | 0.96 |
| P-value (Spearman test) | $1.8 \times 10^{-55}$ | $6.8 \times 10^{-54}$ | $2.3 \times 10^{-16}$ | $1.6 \times 10^{-10}$ | $1.4 \times 10^{-45}$ | $4.0 \times 10^{-56}$ |

In total, 572 incorrect model outputs were linked to options annotated as low, moderate, or high severity according to predefined clinical impact criteria, which are detailed in the Methods section. Overall percent agreement among raters was 55%. Mean observed agreement was $\bar{P} = 0.35$, expected agreement under chance was $\bar{P}_e = 0.34$, yielding a Fleiss' κ of 0.02, indicating minimal agreement beyond chance despite moderate raw agreement. Severity was not dominated by low-risk errors: low-severity errors comprised 0.28, moderate 0.42, and high 0.31 of cases. Thus, 72% of incorrect outputs fell into moderate or high severity categories. The near-zero κ indicates that although models may fail on the same questions, the clinical implications of those failures are not tightly clustered within a single severity category. Clinical severity was evaluated independently of entropy, consensus strength, and robustness. Accordingly, severity represents an orthogonal safety-relevant axis: improvements in stability or robustness do not eliminate moderate- and high-severity error modes.

## 2.8. Agentic reasoning reshapes collective decision structure without eliminating coordinated error

Across analyses, a consistent structural pattern emerges. Agentic retrieval-augmented reasoning reduces inter-model dispersion, strengthens majority consensus, and increases robustness of correctness at the population level. These effects indicate that shared structured retrieval aligns heterogeneous models toward more concentrated and more reproducible decisions.

At the same time, improvements are not uniform. Although agreement amplification occurs more often in settings where the majority is correct, agentic reasoning can also concentrate models around incorrect answers. Robustness gains dominate overall, yet a small subset of questions exhibits coordinated decreases, reflecting synchronized failure rather than isolated model errors. Consensus strength and robust correctness remain strongly aligned on average under both inference strategies. Correct majorities consistently show higher agreement than incorrect majorities, yet high-consensus and low-robustness cases persist. Strong agreement therefore does not guarantee broadly shared correctness. Verbosity does not function as a reliable signal of validity, with largely overlapping length distributions between correct and incorrect responses. Clinical severity analysis further shows that incorrect outputs are frequently associated with moderate or high potential impact, and severity profiles remain heterogeneous. Improvements in stability and robustness therefore do not eliminate clinically consequential error modes.

Single-model accuracy changes were heterogeneous across the 34 LLMs (**Table 1**). Several smaller and mid-sized models showed statistically significant gains under agentic inference, whereas many higher-performing models changed little. Thus, population-level improvements in dispersion and robustness do not imply uniform gains for every individual model, underscoring the value of evaluating collective behavior beyond per-model accuracy alone.

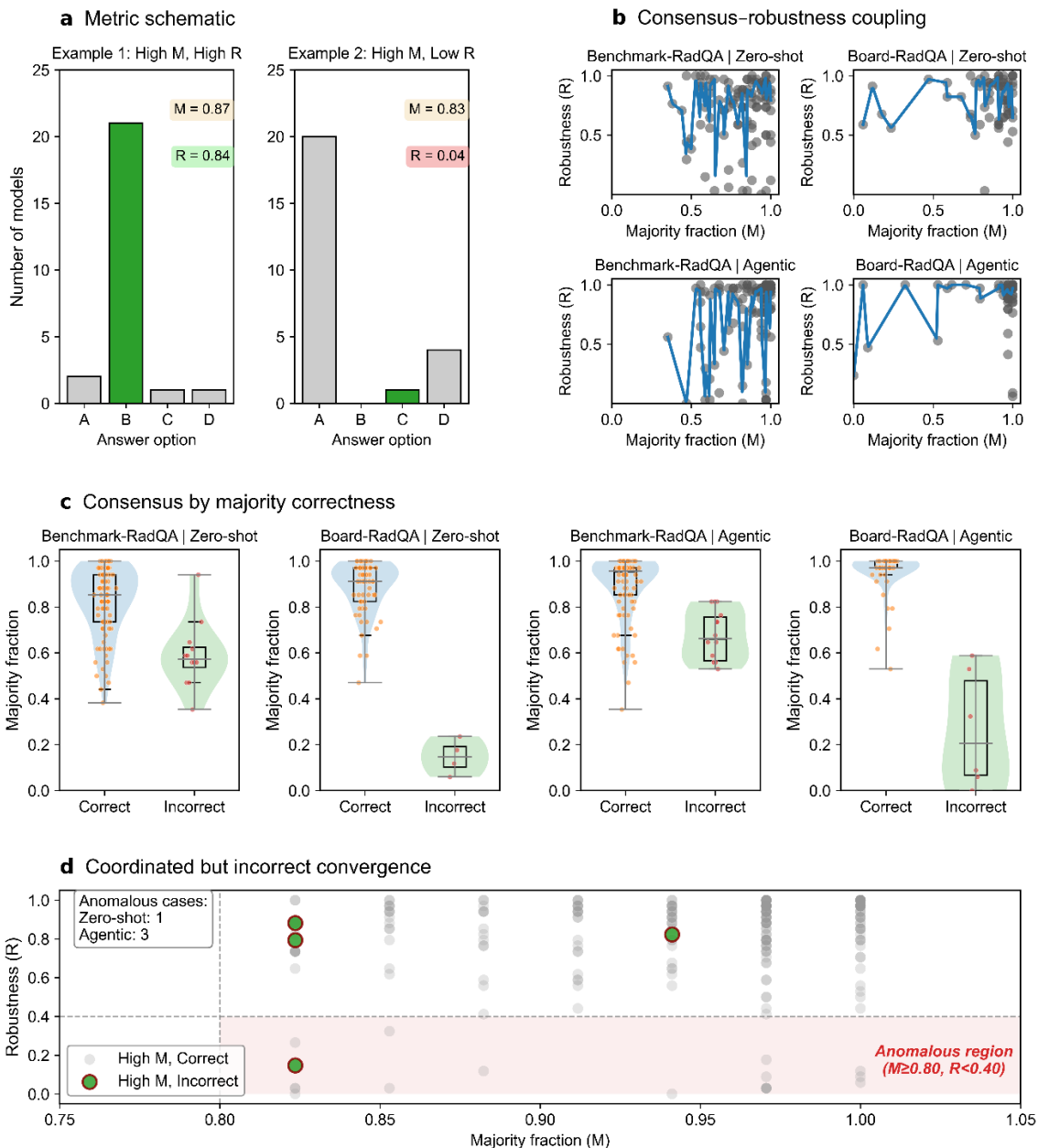

**Figure 5:** Coupling between consensus strength and robust correctness and coordinated incorrect convergence. Consensus strength was quantified as the majority fraction (M), and robust correctness as the fraction of models selecting the ground-truth answer (R), both computed per question. Results are shown separately for Benchmark-RadQA and Board-RadQA under zero-shot and agentic inference. **a** Metric schematic illustrating representative scenarios. Example 1 shows high consensus and high robustness (high M, high R), where most models agree and most are correct. Example 2 shows high consensus but low robustness (high M, low R), where many models agree on the same answer but few select the correct option. **b** Scatter plots of robustness (R) vs. majority fraction (M) at the per-question level for each dataset and inference condition. Each point represents one question. Lines illustrate monotonic trends corresponding to Spearman rank correlations, quantifying the strength of coupling between agreement and correctness. **c** Distribution of majority fractions stratified by whether the majority decision was correct or incorrect. Violin plots with embedded boxplots display per-question majority fractions for correct-majority and incorrect-majority cases within each dataset and inference strategy. **d** Identification of coordinated but incorrect convergence. Points denote per-question (M, R) pairs, with the shaded region highlighting the predefined anomalous zone (M ≥ 0.8 and R < 0.4). High-consensus incorrect cases are marked, illustrating instances in which strong inter-model agreement coexisted with low robustness of correctness.

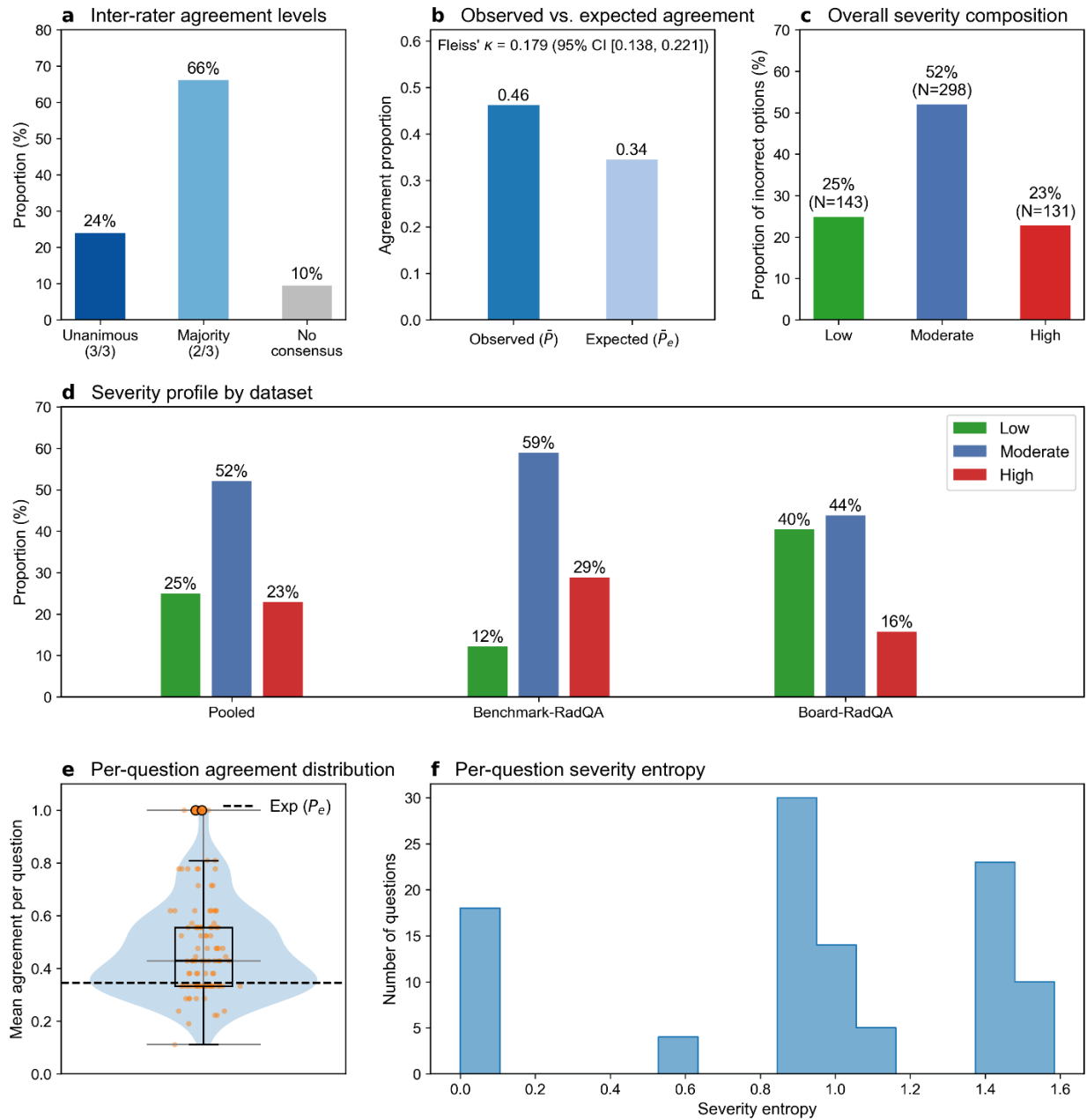

**Figure 6:** Clinical severity distribution and inter-rater agreement for incorrect answer options. Clinical severity of incorrect multiple-choice options was independently assessed by three radiologists and categorized as low, moderate, or high anticipated clinical impact. **a** Inter-rater agreement levels across all annotated options. Bars show the proportion of options with unanimous agreement (3/3), majority agreement (2/3), or no consensus among raters. **b** Observed vs. expected agreement proportions used to compute Fleiss' κ. The observed mean agreement exceeded the expected agreement under chance, yielding a κ value of 0.02. **c** Overall severity composition across all annotated incorrect options (n = 572). Bars show the proportion and absolute counts of low-, moderate-, and high-severity categories. **d** Severity profile stratified by dataset (pooled, Benchmark-RadQA, Board-RadQA), displaying the percentage of incorrect options assigned to each severity level within each subset. **e** Distribution of mean per-question agreement across raters. The violin plot with embedded boxplot shows variability in agreement at the question level; the dashed horizontal line indicates the expected agreement under chance. **f** Distribution of per-question severity entropy, quantifying heterogeneity of severity labels across incorrect options within each question. Higher entropy values reflect greater dispersion of severity categories, indicating that clinical impact of errors is not concentrated in a single severity level.

# 3. Discussion

This study evaluated how a standardized agentic retrieval-augmented[13] reasoning pipeline[8] reshapes collective decision behavior across a heterogeneous panel of 34 LLMs on 169 radiology multiple-choice questions spanning two datasets[8,23]. Rather than focusing on single-model accuracy, we analyzed reliability under model variability by decomposing collective behavior into decision stability (entropy), consensus strength (majority fraction), robustness of correctness (fraction of models correct), the coupling between consensus and correctness, the relationship between verbosity and correctness, and clinically annotated error severity. Three radiologists independently assessed the clinical severity of incorrect answer options to provide a safety-relevant perspective that is not captured by correctness alone.

Agentic inference was associated with lower inter-model dispersion, reflected by a significant reduction in entropy in the pooled analysis. This pattern suggests that shared structured retrieval context tends to align heterogeneous models toward fewer distinct answer modes. Importantly, lower entropy reflects stronger coordination among models but does not imply higher validity[27]. Agreement strength also increased under agentic inference, with higher majority fractions indicating more concentrated collective decisions. However, agreement amplification was not exclusively correctness-preserving. Although most increases in consensus occurred in questions with correct majorities, some questions showed increased agreement around incorrect answers. These observations highlight that stability and consensus represent structural properties of collective behavior rather than direct indicators of correctness.

Robustness of correctness increased under agentic inference, indicating that a larger fraction of models reached the correct answer for many questions. Because robustness captures reproducibility across model variability rather than average performance, this shift is relevant for deployment settings in which model identity or configuration may change. At the same time, improvements were not uniform. A small subset of questions exhibited pronounced robustness decreases, including rare collapse events in which many models shifted away from the correct answer under agentic inference. These coordinated failures illustrate that shared reasoning context can occasionally synchronize errors across models, reducing the protective value of model diversity[16,18]. In this sense, agentic reasoning appears to modify the distribution of errors, combining broader gains in cross-model correctness with residual tail-risk behavior.

The relationship between consensus and correctness further clarifies this structure. Consensus strength and robustness were strongly correlated across questions under both inference strategies, indicating that stronger agreement generally coincided with higher fractions of correct models. Correct majority decisions also showed substantially higher agreement than incorrect majorities. Nevertheless, high agreement did not guarantee correctness. A small number of high-consensus, low-robustness cases occurred under both inference conditions, demonstrating that coordinated incorrect convergence can arise even when models appear highly aligned. Agreement therefore remains an informative but imperfect signal of reliability[30].

Verbosity behaved similarly as a weak indicator of validity[31]. Response lengths for correct and incorrect outputs largely overlapped, and effect sizes were negligible. Under agentic inference, no meaningful relationship between verbosity and correctness was observed. These findings suggest that response length or explanatory detail should not be interpreted as a reliable proxy for correctness, particularly in pipelines that systematically increase output length through structured reasoning steps[32,33].

Clinical severity analysis provided an additional safety-relevant perspective. Incorrect model outputs were frequently associated with moderate or high potential clinical impact t[34], and inter-rater agreement[29] beyond chance was low despite moderate raw agreement. This likely reflects the contextual complexity of judging downstream clinical consequences. Because severity annotations were evaluated independently of correctness, entropy, consensus, and robustness, they represent an orthogonal dimension of model behavior. The persistence of moderate- and high-severity errors indicates that structural improvements in collective correctness do not eliminate clinically consequential failure modes. Reliability metrics describe how often and how consistently models are correct, whereas severity characterizes the potential consequences of residual errors.

Our study has limitations. First, the evaluation is based on 169 curated multiple-choice questions across two datasets. This paired[35], per-question design enables controlled comparison of inference conditions and precise estimation of structural metrics such as entropy, robustness, and consensus coupling. However, the total sample size limits statistical power for fine-grained subgroup analyses, including question-type stratification, pathology-specific effects, or retrieval-pattern sensitivity[36]. In addition, multiple-choice questions abstract away from the open-ended reasoning typical of real clinical reporting. Larger and more diverse datasets, including broader pathology coverage and varying difficulty levels, would allow more granular modeling of collapse-prone scenarios and stronger statistical certainty for tail-risk analyses. Second, the evaluation is text-only and does not incorporate imaging data or multimodal clinical context. Real-world radiology workflows integrate images, reports, prior examinations, and evolving clinical information[1,22]. The present design isolates reasoning over textual knowledge, which strengthens internal validity but constrains external generalizability[37]. Future research should extend the collective-behavior framework to multimodal settings, assessing whether the observed structural shifts under agentic reasoning persist when visual features and heterogeneous inputs are introduced. Prospective validation embedded in reporting workflows would be necessary to evaluate practical utility and user impact[38]. Third, Third, answer adjudication relied on a structured binary correct or incorrect classification with rule-based extraction of the final explicitly stated answer option when multiple options were mentioned. Although this approach ensured consistency and reproducibility, it assumes that the last explicitly stated option reflects the model's intended final decision. In rare cases of ambiguous phrasing or complex answer formulations, this rule-based determination may imperfectly capture the model's true intent. More nuanced adjudication procedures incorporating structured parsing or independent human verification could further reduce potential misclassification risk. Fourth, the standardized[8] agentic pipeline is standardized and fixed across all models to isolate inter-model variability under identical structured context[35]. This design choice ensures controlled comparisons but also constrains architectural diversity in retrieval and report synthesis. Alternative retrieval sources, ranking strategies, prompt templates, or evidence synthesis formats may produce different stability, robustness, and collapse patterns. Systematic ablation studies varying retrieval depth, evidence diversity, or report structure would help disentangle which components drive robustness gains vs. synchronized failures. Fifth, all models received identical retrieved context per question. While this design isolates how different models respond to shared evidence, it may amplify correlated errors when retrieval is misleading or incomplete[18]. The observed robustness collapses likely reflect this shared-context effect. Future systems could explore retrieval diversity, ensemble retrieval strategies, or evidence-quality indicators that detect low-confidence or conflicting context before it is broadcast to all models. Introducing controlled heterogeneity at the retrieval stage may help preserve robustness gains while reducing synchronized failure risk. Sixth, clinical severity labeling is inherently subjective and context dependent. Although three blinded clinicians independently annotated incorrect options, chance-corrected agreement was low, reflecting the nuanced and scenario-specific nature of harm assessment[39]. Certain question types, such as differential diagnoses or technical items, do not map directly onto immediate clinical consequence. Their downstream impact depends on probability of harm, opportunity for correction, and whether an error would propagate into management decisions. Severity judgments may therefore implicitly combine consequence magnitude with assumptions about detectability and reversibility[40]. While this does not invalidate the aggregated severity distribution, it highlights conceptual ambiguity in consequence modeling. Larger annotation panels, structured adjudication procedures, or explicitly probabilistic severity frameworks may improve reliability and transparency[34].

In summary, agentic retrieval-augmented reasoning appears to modify the collective structure of radiology question answering under model variability. In this setting, it was associated with reduced inter-model dispersion, stronger majority consensus, and higher robustness of correctness across the model panel. At the same time, consensus remained an imperfect indicator of correctness, and coordinated incorrect convergence as well as rare robustness collapses were still observed. Improvements in collective robustness therefore coexist with residual tail-risk behavior and clinically consequential error modes. These findings suggest that evaluating agentic systems solely through average accuracy or agreement may be insufficient. Complementary analyses of stability, cross-model robustness, and the potential clinical impact of residual errors may provide a more complete view of reliability under deployment variability.

# 4. Methods

## 4.1. Ethics statement

All procedures followed relevant guidelines and regulations. This study used previously published and expert-curated question datasets and did not involve patients, identifiable personal data, or human subjects. Institutional review board approval and informed consent were therefore not required.

## 4.2. Datasets

We evaluated models on 169 multiple-choice radiology questions drawn from two expert-curated datasets: a Benchmark Radiology QA dataset (Benchmark-RadQA; n = 104), sourced from the RadioRAG study[23], and a board-style Radiology QA dataset (Board-RadQA; n = 65), sourced from the RaR study[8].

### *4.2.1. Benchmark-RadQA (n = 104; RadioRAG-derived)*

Benchmark-RadQA[23] is a benchmark-style radiology question answering dataset constructed by combining two components from the RadioRAG study: RSNA-RadioQA[23] and ExtendedQA[23] **(Supplementary Note 3)**. RSNA-RadioQA was curated from 80 peer-reviewed cases in the Radiological Society of North America (RSNA) Case Collection. Questions were created from the clinical history and image characteristics described in the case documentation and figure captions, while images themselves were not included to enforce a text-only evaluation setting. During curation, differential diagnoses explicitly listed by original case authors were excluded to reduce leakage of the correct answer. The dataset spans 18 radiology subspecialties (with at least five questions per subspecialty in the source dataset), reflecting a broad coverage of diagnostic radiology topics **(Supplementary Table S1)**. ExtendedQA was developed to probe generalization and reduce the risk of data contamination in evaluation. It consists of 24 radiology questions developed and validated by board-certified radiologists with substantial diagnostic radiology experience (5–14 years)[23]. These questions were designed to reflect realistic clinical diagnostic scenarios that were not previously available online in the same form as standard case collections. RSNA-RadioQA is derived from a publicly accessible case collection; therefore, some degree of training-data cannot be fully excluded for any evaluated model family. ExtendedQA and Board-RadQA were included to probe generalization under reduced online availability of question formulations.

Because ExtendedQA[23] was originally open-ended, we used the standardized preprocessing described in the RaR paper to harmonize format and scoring across all 104 Benchmark-RadQA questions. Specifically: (i) ExtendedQA questions were converted to multiple-choice format while preserving the original stem and correct answer, (ii) to standardize evaluation across both RSNA-RadioQA and ExtendedQA, three clinically plausible distractors were generated for every question to produce four answer choices per item (one correct answer plus three distractors), and (iii) all distractors were reviewed and curated by a board-certified radiologist to ensure plausibility, difficulty, and absence of implausible or misleading options. Distractor drafts were generated using OpenAI's GPT-4o and o3 models, but these models did not determine ground truth; they were used only to propose candidate distractors that were then filtered and finalized through expert review. The resulting Benchmark-RadQA dataset used here therefore contains 104 multiple-choice questions with four options per question (A–D).

### *4.2.2. Board-RadQA (n = 65; board-style, RaR-derived)*

Board-RadQA[8] is a set of 65 board-style radiology questions aligned with German radiology board examination domains. Questions were developed and validated by board-certified radiologists with 9–10 years of experience and reflect core diagnostic knowledge emphasized in structured radiology training. According to the RaR study[8], these questions and their formulations are not available in online case collections or known LLM training corpora, reducing the likelihood of training-data overlap. Board-RadQA questions were formatted as five-option multiple-choice items with a single reference-standard correct answer. The dataset is publicly available for research use as reported in the RaR study[8].

Across both datasets, identical question text, answer options, and ground-truth labels were used across all models and inference conditions.

## 4.3. Model panel

We evaluated a fixed panel of 34 heterogeneous language models spanning a wide range of parameter scales, training paradigms, and access models. The panel was designed to approximate realistic deployment heterogeneity rather than to rank individual systems. The evaluated models included proprietary and open-weight systems from multiple families. Concretely, the panel comprised: Claude-Sonnet-4.6, Gemini-3.1-Pro (Preview)[41], MiniMax-M2.5[42], GLM-5[43], LFM2.5-1.2B-Thinking, Kimi-K2.5[44], Palmyra-X5, MiMo-V2-Flash[45], Ministral-8B, Mistral Large, Llama3.3-8B[46,47], Llama3.3-70B[46,47], Llama3-Med42-8B[48], Llama3-Med42-70B[48], Llama4-Scout-16E[33], DeepSeek R1-70B[49], DeepSeek-R1[49], DeepSeek-V3[50], Qwen 2.5-0.5B[51], Qwen 2.5-3B[51], Qwen 2.5-7B[51], Qwen 2.5-14B[51], Qwen 2.5-70B[51], Qwen 3-8B[52], Qwen 3-235B[52], GPT-3.5-turbo, GPT-4-turbo[53], o3, GPT-5[54], GPT-5.2, MedGemma-4B-it[55], MedGemma-27B-text-it[55], Gemma-3-4B-it[56,57], and Gemma-3-27B-it[56,57]. These models range from sub-billion to very large-scale architectures and include general-purpose, instruction-tuned, and medically adapted variants. Detailed specifications, access modes, and version identifiers are provided in **Supplementary Table S2**. All models were run using the lowest-variance decoding settings available for each API (e.g., minimum supported temperature; top-p set to 1 or disabled when applicable). For OpenAI reasoning models (e.g., o3), sampling parameters such as temperature/top-p are not user-configurable, so we used the model defaults. We generated one response per question per inference condition[8].

## 4.4. Inference conditions

Each model answered every question under two inference strategies: zero-shot inference and agentic retrieval-augmented reasoning (**Supplementary Notes 4** and **5**).

In the zero-shot condition, models received only the question stem and answer options and were instructed to select the single best answer. No external retrieval, tools, or iterative reasoning scaffolds were provided. A standardized prompt template was used across models and datasets, instructing the model to act as a medical expert and to select exactly one option. The exact prompt templates are provided in **Supplementary Note 5**.

In the agentic condition, models received additional structured context generated by a fixed retrieval-augmented[13,23] orchestration pipeline adapted from prior radiology-focused retrieval and reasoning frameworks. The pipeline comprised three sequential stages: (i) automated extraction and abstraction of key diagnostic concepts from the question stem, (ii) multi-step evidence retrieval restricted to Radiopaedia.org[26], a peer-reviewed radiology knowledge base, and (iii) synthesis of the retrieved content into a standardized structured report generated by a single, fixed orchestrator model. This report is then provided to the target model as additional context before answer selection.

The orchestration process was held fixed across all evaluated models. All models received identical retrieved context for a given question, and the final answer was always generated by the evaluated model itself. The orchestration engine thus functioned only as a context-construction mechanism and not as a decision-maker. This design isolates how different models use the same external evidence rather than comparing retrieval or planning abilities across models[8]. Prompt templates and representative examples of retrieved-context formatting are provided in **Supplementary Notes 4** and **5** to support reproducibility.

## 4.5. Answer adjudication

Responses were first evaluated against the reference-standard correct option for each question using a structured adjudication procedure with binary correct or incorrect classification. For responses classified as correct, scoring was based directly on the reference-standard option. For responses classified as incorrect, the generated text was automatically reviewed using rule-based matching to identify the final explicitly stated answer option, which was treated as the model's definitive selection. This procedure ensured a consistent and reproducible determination of the model's intended answer, including cases in which multiple options were mentioned within a single response[8].

## 4.6. Evaluation framework and experimental design

Our evaluation targets collective behavior across models under model variability. Analyses were defined at the per-question level and paired across inference conditions. For each question $q$, each of the $N = 34$ models produced one response under zero-shot and one under agentic inference, yielding 68 responses per question and 11,492 total responses across 169 questions. Unless otherwise stated, the question is the statistical unit of analysis and comparisons are paired[35] by question.

### 4.6.1. Inter-model decision stability

Inter-model decision stability was used to quantify how consistently a heterogeneous panel of language models converged on the same answer when presented with an identical radiology question under a fixed inference strategy. The underlying rationale is that, when multiple independent models are exposed to the same task, the dispersion of their discrete decisions provides a measure of collective stability that is distinct from correctness.

For each radiology question and each inference condition, we collected one final answer from each of the 34 models. Answers were restricted to the predefined multiple-choice options of the question, yielding a finite categorical outcome space. For a given question under a given method, the distribution of model decisions over the available options defines an empirical categorical distribution. If $n(o)$ denotes the number of models selecting option $o$ and $N = 34$ denotes the total number of models, the empirical probability of option $o$ is defined as $p(o) = \frac{n(o)}{N}$.

Decision stability was quantified using Shannon entropy[27], defined as:

$$H = -\sum_{o \in O} p(o) \log p(o), \tag{1}$$

where $O$ denotes the set of available answer options for that question. By convention, terms with $p(o) = 0$ contribute zero to the sum. Entropy equals zero when all models select the same option and increases as responses become more evenly distributed across options. In this framework, entropy captures dispersion of decisions without reference to their correctness. A single entropy value was computed for each question under each inference condition. This produced paired entropy measurements per question, one for zero-shot inference and one for agentic retrieval-augmented reasoning. Inter-model stability was therefore operationalized as a question-level property that could be directly compared across inference conditions while holding the question constant.

To characterize how the inference strategy influences collective decision structure, paired differences in entropy between conditions were computed at the question level. These paired values form the basis for distributional summaries. Importantly, this stability metric is correctness-agnostic and is intended to capture coordination structure rather than validity[15,16]. This separation allows subsequent analyses to disentangle agreement, correctness, and robustness as distinct dimensions of model behavior.

### *4.6.2. Majority decision behavior*

Majority decision behavior was analyzed to characterize how strongly a panel of heterogeneous models converges on a single answer and whether such convergence aligns with the reference standard[58]. Whereas entropy captures the full dispersion of responses, the majority-based analysis focuses on the dominant collective decision and the strength with which it is supported[27].

For each radiology question under each inference condition, we considered the set of final answer options selected by the 34 models. Let $n(o)$ denote the number of models selecting option $o$ from the option set $O$. The majority option $o^*$ is defined as the option with the highest frequency among model responses, $o^* = \arg\max_{o \in O} n(o)$. The strength of consensus was quantified by the majority fraction, $M = \frac{\max_{o \in O} n(o)}{N}$, where $N = 34$ is the number of models. This quantity represents the proportion of models supporting the most common answer and lies between the reciprocal of the number of options and 1. Higher values indicate stronger concentration of decisions on a single option.

To relate consensus to validity, the majority decision was compared with the ground-truth answer for that question. Majority correctness was treated as a binary property indicating whether the majority option coincided with the reference standard. This labeling does not alter the definition of consensus itself but allows subsequent analyses to examine how agreement and correctness interact. Each question yielded two majority fractions, one under zero-shot inference and one under agentic retrieval-augmented reasoning. The question-level change in consensus strength was defined as the paired difference $\Delta M = M_{\text{agentic}} - M_{\text{zero-shot}}$. This paired formulation ensures that differences are evaluated within the same question, controlling for variation in topic and difficulty[35]. For interpretive analyses, questions were grouped according to whether consensus strength increased, decreased, or remained unchanged between conditions, and whether the majority decision under the agentic condition was correct or incorrect. These categorizations serve to describe how shifts in collective agreement relate to decision validity without assuming that consensus is itself a reliability metric.

### *4.6.3. Robustness of correctness across models*

Robustness of correctness was evaluated to quantify how consistently a question is answered correctly across a heterogeneous model panel and to what extent correctness depends on model choice[18]. Unlike accuracy (defined per model as the fraction of questions answered correctly), robustness is defined per question as the fraction of models that answer correctly, capturing cross-model reproducibility of correct decisions[11,15]. For each question under each inference condition, the correctness of each model response was determined relative to the reference standard. Let $c_i \in \{0,1\}$ denote the correctness indicator for model $i$, where $c_i = 1$ indicates a correct answer and $c_i = 0$ otherwise. With $N = 34$ models, robustness of correctness for a given question and method is defined as:

$$R = \frac{1}{N} \sum_{i=1}^{N} c_i, \tag{2}$$

which corresponds to the empirical probability that a randomly selected model from the panel answers the question correctly. This quantity lies in the unit interval, with higher values indicating that correctness is reproducible across many independent model instances rather than driven by a small subset.

To facilitate interpretable summaries of reliability regimes, robustness values were additionally stratified into three ordinal categories representing low, intermediate, and high cross-model consistency. These categories were defined by fixed thresholds on $R$ and used exclusively for descriptive transition analyses, whereas all statistical testing was conducted on the continuous robustness values to avoid discretization artifacts[59]. Each question yielded two robustness values, one under zero-shot inference and one under agentic retrieval-augmented reasoning. The question-level change in robustness was defined as the paired difference $\Delta R = R_{\text{agentic}} - R_{\text{zero-shot}}$, which isolates the effect of the inference strategy while holding the question constant[35]. Positive values indicate that a larger fraction of models reached the correct answer under the agentic condition, whereas negative values indicate reduced cross-model consistency of correctness.

For descriptive transition analyses, questions were further characterized according to whether their robustness category increased, decreased, or remained stable between inference conditions. This transition view is intended to reveal structural shifts in reliability regimes, such as movement from fragile to consistently correct behavior or, conversely, coordinated degradations. Importantly, robustness is interpreted as a measure of cross-model stability of correctness and not as a proxy for clinical validity or task difficulty.

### *4.6.4. Output verbosity as a confidence proxy*

To examine whether commonly exposed verbosity signals relate to decision validity, we analyzed the association between response length and correctness at the level of individual model outputs. The underlying question is whether longer or more detailed responses systematically correspond to higher likelihood of correctness, which would make verbosity a plausible proxy for confidence[19,20]. Each model response produced under each inference condition constituted one observation. For every response, correctness was determined relative to the reference standard, yielding a binary indicator $c \in \{0,1\}$. Two quantitative verbosity measures were extracted from the textual outputs. Reasoning length was defined as the number of tokens in the model's explanatory reasoning segment when present, and summary length as the number of tokens in the final answer or summary segment[60]. Token counts were computed using the OpenAI's tiktoken tokenizer, a fast byte-pair encoding with the $cl100k\_base$ encoding. Formally, for a given verbosity measure $L$ and inference method $m$, we consider the conditional distributions $L \mid c = 1, m$ and $L \mid c = 0, m$. These represent the length distributions for correct and incorrect responses, respectively, under the same inference condition. The analysis tests whether these distributions differ in location or spread, without assuming any specific parametric form.

All comparisons were performed within inference condition, so that zero-shot responses were only compared to other zero-shot responses and agentic responses only to agentic responses. This isolates the relationship between verbosity and correctness from systematic length differences induced by the inference pipeline itself. Throughout, verbosity measures are treated strictly as descriptive behavioral signals. They are not interpreted as calibrated confidence scores[17], and no assumption is made that longer outputs imply greater reliability or safety[19,20].

### *4.6.5. Coupling between consensus strength and robust correctness*

To examine whether stronger inter-model agreement corresponds to more reliable correctness across models, we analyzed the coupling between consensus strength and robustness at the per-question level[18,61]. This analysis integrates two previously defined quantities: the majority fraction, which captures how concentrated model answers are on a single option, and the robustness score, which captures how many models independently arrive at the correct answer. The goal is to determine whether these two dimensions track each other or can diverge. For each question $q$ and inference method $m$, consensus strength was represented by the majority fraction $M_{q,m}$, defined as the proportion of models selecting the most frequent answer option, and robust correctness was represented by the robustness score $R_{q,m}$, defined as the fraction of models that selected the ground-truth option. Both quantities lie in the unit interval [0, 1] but capture different aspects of collective behavior: $M_{q,m}$ is agnostic to correctness, whereas $R_{q,m}$ is correctness-anchored.

The association between consensus and robustness was quantified separately for each inference method using Spearman's rank correlation coefficient[62]. Spearman correlation was chosen because it evaluates monotonic association without assuming linearity or normality and is appropriate for bounded, non-Gaussian variables. Formally, for each method $m$, the correlation $\rho_m$ was computed over paired observations $(M_{q,m}, R_{q,m})$ across questions. This analysis assesses whether questions with stronger consensus also tend to exhibit higher fractions of correct models.

To further test whether higher consensus preferentially occurs on correct collective decisions, we compared the distribution of majority fractions between questions where the majority decision matched the reference standard and those where it did not. This comparison isolates whether consensus strength itself is systematically aligned with correctness at the majority-vote level rather than at the individual-model level.

Finally, we explicitly characterized consensus-related failure modes by identifying questions for which strong agreement coexisted with weak correctness. These were defined a priori as questions satisfying:

$$M_{q,m} \geq 0.8 \quad \text{and} \quad R_{q,m} < 0.4, \tag{3}$$

meaning that a large majority of models agreed on an answer while fewer than 40% of models selected the correct one. Such cases represent coordinated but incorrect convergence. Their frequency was summarized descriptively for each inference method. This analysis is intended to reveal structural patterns in collective behavior rather than to define safety thresholds or calibration properties.

### *4.6.6. Clinical severity assessment of incorrect decisions*

To evaluate the potential clinical risk associated with incorrect model decisions, we conducted an independent expert severity assessment performed by two board-certified radiologists (L.A., T.T.N.) and one final-year radiology resident (F.B.O.), with 10, 8, and 5 years of clinical experience in diagnostic and interventional radiology across subspecialties. This component was designed to quantify the clinical consequence of incorrect answers rather than to reassess model accuracy. The unit of annotation was the incorrect answer option. For each question in the combined dataset of 169 radiology questions, all incorrect multiple-choice options were evaluated. Radiologists were fully blinded to model outputs, model identities, inference strategies, and all aggregate statistics derived from model behavior, including agreement levels, entropy, consensus measures, and robustness scores. They were also not informed how frequently any option had been selected by models. This blinding ensured that severity assessments reflected independent clinical judgment rather than perceptions of model performance or consensus[63].

Radiologists were instructed to judge the likely clinical consequence if a clinician were to select a given incorrect option as the final diagnostic decision. Severity was defined in terms of potential impact on patient management and outcomes[64]. A high-severity error corresponded to an incorrect diagnosis that could plausibly lead to substantial patient harm, major diagnostic delay, or inappropriate management with meaningful clinical risk. A moderate-severity error corresponded to an incorrect diagnosis that could lead to diagnostic delay or suboptimal management with limited or reversible clinical impact. A low-severity error corresponded to an incorrect diagnosis unlikely to meaningfully affect patient outcomes or management[34]. Correct options were not assigned severity labels and served only to determine which options were eligible for annotation.

Inter-rater reliability across the three radiologists was quantified using percent agreement and Fleiss' κ, which extends chance-corrected agreement measures to multiple raters[65]. Fleiss' κ was computed as $\kappa = \frac{\bar{P} - \bar{P}_e}{1 - \bar{P}_e}$, where $\bar{P}$ denotes the observed mean agreement across raters and $\bar{P}_e$ denotes the expected agreement under chance. This metric was selected because severity labels are categorical and independently assigned by more than two raters.

For downstream analyses, option-level severity labels were aggregated using majority agreement across the three radiologists. In cases without a strict majority, the median severity on the ordinal scale from low to high was assigned according to a prespecified rule. These aggregated severity labels were subsequently linked to model-level failure patterns, including incorrect majority decisions, high-consensus incorrect decisions, and robustness-collapse cases.

Severity labels were used exclusively for descriptive and stratified analyses of safety-relevant failure modes. They were not incorporated into correctness scoring and did not influence entropy, consensus, or robustness calculations. In this way, clinical severity was treated as an independent safety-relevant dimension that complements, but does not redefine, statistical performance metrics. This design allows a clear separation between the frequency of errors, the consistency of errors across models, and the potential clinical consequence of those errors[9].

## 4.7. Statistical analysis

MF, JS, SN, and STA performed the evaluations and statistical analyses between January and March 2026. Statistical analyses and data visualization were conducted using Microsoft Excel (Microsoft® Excel® for Microsoft 365 MSO (v2602 Build 16.0.19725.20126) 64-bit) and Python (Python v3.11.7 | packaged by Anaconda, Inc.) Specifically, the primary calculations for inter-model decision stability (entropy), majority fraction, and clinical severity reported in the text and tables were performed using Microsoft Excel. To ensure accuracy and reproducibility, these calculations were independently cross validated. For the remaining analyses, the following packages were used: NumPy (v1.24.0) and Pandas (v2.0.0) for data handling, SciPy (v1.10.0) for statistical analysis, Matplotlib (v3.7.0) and Seaborn (v0.12.0) for visualization, Plotly (v5.0.0) with Kaleido (v0.2.1) for figure export, OpenPyXL (v3.0.0) for Excel file handling, and TikToken (v0.5.0) for tokenization-based measurement of response length. Single-model accuracy results (**Table 1**) were estimated using bootstrapping with 1,000 resamples to compute means, standard deviations, and 95% confidence intervals (CI)[66] across the pooled dataset of 169 questions. A strictly paired design ensured identical resamples across conditions[67]. To evaluate statistical significance for model-level comparisons between zero-shot and agentic retrieval-augmented methods, exact McNemar's test[68] based on the binomial distribution was applied separately to each model using paired question-level outcomes. P-values were adjusted for multiple comparisons using the false discovery rate[69], with a significance threshold of 0.05.

All other hypothesis tests were two-sided with a significance threshold of 0.05, and no formal adjustment for multiple comparisons was applied, and results should be interpreted in the context of multiple hypothesis testing. The primary unit of analysis was the question. Core comparisons between inference strategies used a paired[35] per-question design, pairing zero-shot and agentic results for the same question. Because several metrics are bounded and exhibited heterogeneous distributions, we emphasize distributional summaries and effect sizes alongside p-values. To mitigate type I error inflation across multiple endpoints, we prespecified a hierarchy of outcomes. The primary outcomes were (i) inter-model decision stability (entropy) and (ii) robustness of correctness. Secondary outcomes included majority fraction, consensus-robustness coupling, verbosity-correctness associations, and clinical severity distributions. Secondary analyses are interpreted as exploratory and hypothesis-generating.

For inter-model decision stability (entropy $H$), majority fraction ($M$), and robustness of correctness ($R$), we assessed systematic differences between agentic and zero-shot inference using the Wilcoxon signed-rank test[70] applied to paired per-question values. This non-parametric test was chosen because it does not assume normality and is appropriate for bounded or skewed distributions. For each metric $X \in \{H, M, R\}$, we defined the paired difference at the question level as $\Delta X_q = X_{q,\text{agentic}} - X_{q,\text{zero-shot}}$.

We report medians, interquartile ranges (IQRs), and distributional characteristics for both raw values and paired differences. In accordance with standard signed-rank procedures, questions with exactly zero paired difference were excluded from the Wilcoxon test statistic; the number of non-zero pairs is reported where relevant. Effect sizes for paired tests are reported as rank-biserial correlations $r$[71].

For comparisons between independent groups within the same inference method, we used the Mann-Whitney U test[72]. This was applied when comparing verbosity-related proxies (reasoning length and summary length) between correct and incorrect responses, and when comparing majority-fraction values between majority-correct and majority-incorrect questions. For these analyses, we report group medians, IQRs, two-sided p-values, and effect sizes as Cliff's delta[73] (or rank-biserial correlation[71] where applicable). These effect sizes provide interpretable non-parametric measures of stochastic dominance without assuming specific distributional forms[74,75].

To quantify the association between consensus strength and robustness of correctness, we computed Spearman rank correlations between majority fraction ($M$) and robustness ($R$) separately for each inference method. Spearman correlation was selected because it captures monotonic relationships without assuming linearity or normality[62,65].

For categorical outcomes and predefined thresholds, including majority-behavior categories, robustness-bin transitions, and predefined high-consensus ($M \geq 0.8$) and low-robustness ($R < 0.4$) cases, we report counts and proportions relative to the relevant denominator (dataset or method). These summaries are descriptive and are not subjected to additional hypothesis testing unless explicitly stated[76].

For radiologist severity annotations of incorrect answer options, inter-rater reliability[29] will be summarized using percent agreement and, where appropriate, Cohen's κ[77]. Severity labels are analyzed descriptively and subsequently linked to model-level failure subsets, such as incorrect-majority decisions or high-consensus failures[10,22]. These labels are not used to modify correctness scoring or any core quantitative metric[9].

# Data availability

All data analyzed in this study originate from publicly available, expert-curated radiology question-answering datasets. The Benchmark-RadQA dataset (comprising RSNA-RadioQA and ExtendedQA items) is available through the original RadioRAG publication and its associated open resources. The Board-RadQA dataset is publicly available for research use and can be accessed as reported in the RaR study and its supplementary materials. No new patient data were generated or used in this work.

# Code availability

All code required to reproduce the analyses in this study is publicly available. The full evaluation and analysis pipeline used to compute stability, consensus, robustness, and related metrics is available at: https://github.com/minafarajiamiri/stability. This repository contains scripts for data processing, metric computation, and statistical analyses, and is sufficient to reproduce the results reported in this work from model outputs.

Agentic retrieval-augmented inference in this study used a previously described orchestration pipeline. To support transparency and reproducibility, the implementation used for generating retrieval-augmented reports is publicly available at: https://github.com/sopajeta/RaR. In the present study, this pipeline was treated as a fixed inference component and was not modified beyond configuration for dataset inputs. Its availability allows independent reproduction of the agentic condition. The implementation relies on widely used open-source frameworks, including:

- LangChain Open Deep Research: https://github.com/langchain-ai/deep-research
- LangChain (v0.3.25): https://github.com/langchain-ai/langchain
- LangGraph (v0.4.1): https://github.com/langchain-ai/langgraph
- OpenAI Python SDK (v1.77.0): https://platform.openai.com
- SearXNG metasearch engine: https://github.com/searxng/searxng
- Docker (v25.0.2): https://www.docker.com

Locally hosted models were run between July 1 and August 22, 2025. The following open-weight models were evaluated, with sources listed for reproducibility:

- Qwen 2.5-0.5B: https://huggingface.co/Qwen/Qwen2.5-0.5B
- Qwen 2.5-3B: https://huggingface.co/Qwen/Qwen2.5-3B
- Qwen 2.5-7B: https://huggingface.co/Qwen/Qwen2.5-7B
- Qwen 2.5-14B: https://huggingface.co/Qwen/Qwen2.5-14B
- Qwen 2.5-70B: https://huggingface.co/Qwen/Qwen2.5-72B
- Qwen 3-8B: https://huggingface.co/Qwen/Qwen3-8B
- Qwen 3-235B: https://huggingface.co/Qwen/Qwen3-235B-A22B
- Llama 3.3-8B: https://huggingface.co/meta-llama/Meta-Llama-3-8B
- Llama 3.3-70B: https://huggingface.co/meta-llama/Llama-3.3-70B-Instruct
- Llama 3-Med42-70B: https://huggingface.co/m42-health/Llama3-Med42-70B
- Llama 3-Med42-8B: https://huggingface.co/m42-health/Llama3-Med42-8B
- Llama4 Scout 16E: https://huggingface.co/meta-llama/Llama-4-Scout-17B-16E
- Mistral Large: https://huggingface.co/mistralai/Mistral-Large-Instruct-2407
- Ministral 8B: https://huggingface.co/mistralai/Ministral-8B-Instruct-2410
- Gemma-3-4B-it: https://huggingface.co/google/gemma-3-4b-it
- Gemma-3-27B-it: https://huggingface.co/google/gemma-3-27b-it
- Medgemma-4B-it: https://huggingface.co/google/medgemma-4b-it
- Medgemma-27B-text-it: https://huggingface.co/google/medgemma-27b-text-it
- DeepSeek-V3: https://huggingface.co/deepseek-ai/DeepSeek-V3
- DeepSeek-R1: https://huggingface.co/deepseek-ai/DeepSeek-R1
- DeepSeek-R1-70B: https://huggingface.co/deepseek-ai/DeepSeek-R1-Distill-Llama-70B

These models were served using vLLM v0.9.0 (https://github.com/vllm-project/vllm). Tensor parallelism matched the number of GPUs per node; models under 3B parameters were served without tensor parallelism.

OpenAI proprietary models were accessed via official API. The versions used were:

- GPT-5.2 (2025-08-07)
- GPT-5 (2025-08-07)
- O3 (2025-04-16)

- GPT-4-Turbo (2024-04-09)
- GPT-3.5-Turbo (2024-01-25)

The OpenRouter unified API was used to access more recent models:

- google/gemini-3.1-pro-preview
- anthropic/claude-sonnet-4.6
- z-ai/glm-5
- LiquidAI/LFM2.5-1.2B-Thinking
- minimax/minimax-m2.5
- moonshotai/kimi-k2.5
- writer/palmyra-x5
- xiaomi/mimo-v2-flash

# Acknowledgements

FBO received funding from funding from the Clinician Scientist Program of RWTH Aachen University. SW is supported by BayernKI and the Deutsche Forschungsgemeinschaft (DFG) – 440719683. KB received grants from the European Union (101079894), Bayern Innovativ, German Federal Ministry of Education and Research, Max Kade Foundation, and Wilhelm-Sander Foundation. SN was supported by grants from the DFG (NE 2136/3-1, LI3893/6-1, TR 1700/7-1). DT was supported by grants from the DFG (NE 2136/3-1, LI3893/6-1, TR 1700/7-1) and is supported by the German Federal Ministry of Education (TRANSFORM LIVER, 031L0312A; SWAG, 01KD2215B) and the European Union's Horizon Europe and innovation programme (ODELIA [Open Consortium for Decentralized Medical Artificial Intelligence], 101057091).

# Author contributions

The formal analysis and study conceptualization were conducted by MF, JS, and STA. The original draft was written by STA, SA, MF, and JS, and edited by STA and SN. The code was developed by MF and JS. All the experiments were performed by MF and JS, except for clinical severity assessment which was performed by LA, TTN, and FBO. ML, LA, KB, DT, and STA contributed to data collection. SW contributed to LLM-serving infrastructure for model inference. The illustrations were designed by MF, JS, and STA. The statistical analyses were performed by STA, SN, MF, and JS. LA, FBO, TTN, KB, SN and DT provided clinical expertise. JS, SA, ML, SW, KB, DT and STA provided clinical expertise. The study was defined by STA. All authors read the manuscript and agreed to the submission of this paper.

# Declaration of interests

JS is partially employed at Siemens Healthineers, Germany. ML is employed by Generali Deutschland Services GmbH, Germany and is an editorial board at European Radiology Experimental. LA is a trainee editorial board at *Radiology: Artificial Intelligence*. SW is partially employed by DATEV eG, Germany. KB is a trainee editorial board at *Radiology: Artificial Intelligence*. DT received honoraria for lectures by Bayer, GE, Roche, AstraZeneca, and Philips and holds shares in StratifAI GmbH, Germany, and in Synagen GmbH, Germany. STA is an editorial board at *Communications Medicine* and at *European Radiology Experimental*, and a trainee editorial board at *Radiology: Artificial Intelligence*. The other authors do not have any competing interests to disclose.

# Supplementary information

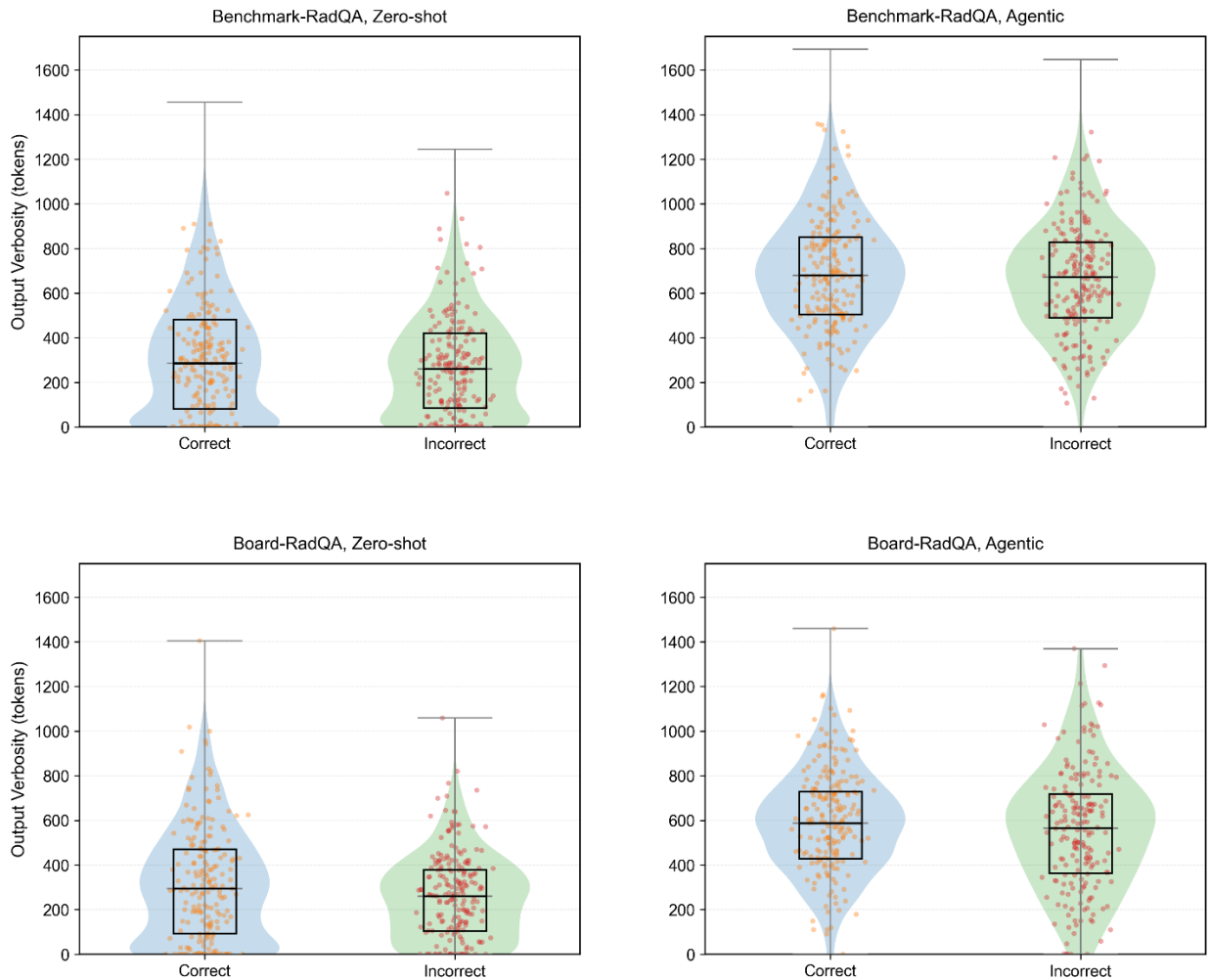

**Supplementary Figure S1:** Output verbosity by correctness under zero-shot and agentic inference. Output verbosity was quantified as total token count per model response. Distributions are shown separately by dataset and inference strategy. Violin plots with embedded boxplots display the distribution of token counts for correct and incorrect responses within each panel. Points represent individual model outputs; boxplots indicate median and interquartile range.

**Supplementary Table S1:** Overview of datasets used to evaluate inter-model decision stability and robustness. Both datasets consist of expert-curated radiology questions spanning multiple subspecialties. Detailed dataset construction procedures are described in the original publications.

| Property | Board-RadQA | Benchmark-RadQA |
|---|---|---|
| Questions [n] | 65 | 104 |
| Question type | Board-style conceptual and clinical | Case-based clinical scenarios |
| Authorship | Board-certified radiologists | Expert-curated, peer-reviewed sources |
| Subspecialty coverage | Multi-subspecialty curriculum-aligned | 18 radiology subspecialties |
| Answer format | 5-option multiple-choice | 4-option multiple-choice |
| Images included | No | No |
| Primary use in prior work | Agentic reasoning evaluation | Online RAG evaluation |
| Original source | RaR study | RadioRAG study |

**Supplementary Table S2:** Specifications of the large language models (LLMs) evaluated in this study. This table summarizes the 34 LLMs assessed under zero-shot prompting and a standard agentic retrieval-augmented generation framework. For each model, we report the parameter count (in billions), category (e.g., instruction-tuned (IT), mixture of experts (MoE), reasoning-optimized, open-weights or proprietary), release date, developer, and maximum context length (in thousands of tokens). All evaluations were conducted between July 1, 2025, and February 25, 2026.

| Model name | Parameters (billions) | Category | Release date | Developer | Context length (thousand tokens) |
|---|---|---|---|---|---|
| Claude-Sonnet-4.6 | Undisclosed | IT, reasoning, proprietary | February 2026 | Anthropic | 1,000 |
| Gemini-3.1-Pro | Undisclosed | Reasoning, proprietary | February 2026 | Google DeepMind | 1,000 |
| MiniMax-M2.5 | 230 | MoE, reasoning, open-weights | February 2026 | MiniMax | 197 |
| GLM-5 | 744 | MoE, open-source | February 2026 | Z.ai | 200 |
| LFM2.5-1.2B-Thinking | 1.17 | IT, open-weights | January 2026 | Liquid AI | 32 |
| Kimi-K2.5 | 1000 | MoE, open-source | January 2026 | Moonshot AI | 256 |
| MiMo-V2-Flash | 309 | MoE, open-source | December 2025 | Xiaomi | 256 |
| Palmyra-X5 | Undisclosed | MoE, proprietary | April 2025 | Writer | 1,000 |
| Llama4-Scout-16E | 17 | IT, open-weights | April 2025 | Meta AI | 10,000 |
| Llama3.3-8B | 8 | IT, open-weights | April 2024 | Meta AI | 8 |
| Llama3.3-70B | 70 | IT, open-weights | April 2024 | Meta AI | 128 |
| Llama3-Med42-8B | 8 | IT, clinically-aligned, open-weights | June 2024 | M42 Health AI Team | 8 |
| Llama3-Med42-70B | 70 | IT, clinically-aligned, open-weights | June 2024 | M42 Health AI Team | 8 |
| DeepSeek R1-70B | 70 | Reasoning, open-source | January 2025 | DeepSeek | 128 |
| DeepSeek-R1 | 671 | Reasoning, open-source | January 2025 | DeepSeek | 128 |
| DeepSeek-V3 | 671 | MoE, open-source | December 2024 | DeepSeek | 128 |
| GPT-5 | Undisclosed | IT, reasoning, proprietary | August 2025 | OpenAI | 128 |
| GPT-5.2 | Undisclosed | IT, reasoning, proprietary | December 2025 | OpenAI | 400 |
| o3 | Undisclosed | Reasoning, proprietary | April 2025 | OpenAI | 200 |
| GPT-3.5-turbo | Undisclosed | IT, proprietary | January 2025 | OpenAI | 16 |
| GPT-4-turbo | Undisclosed | IT, proprietary | April 2024 | OpenAI | 128 |
| Mistral Large | 123 | IT, open-source | July 2024 | Mistral AI | 128 |
| Ministral-8B | 8 | IT, open-source | October 2024 | Mistral AI | 128 |
| MedGemma-4B-it | 4 | Gemma-3-based, multimodal, IT, clinical reasoning, open-weights | July 2025 | Google DeepMind | 128 |
| MedGemma-27B-text-it | 27 | Gemma-3-based, text only, IT, clinical reasoning, open-weights | July 2025 | Google DeepMind | ≥ 128 |
| Gemma-3-4B-it | 4 | IT, open-weights | March 2025 | Google DeepMind | 128 |
| Gemma-3-27B-it | 27 | IT, open-weights | March 2025 | Google DeepMind | 128 |
| Qwen 3-8B | 8 | Reasoning, MoE, open-source | April 2025 | Alibaba Cloud | 32 |
| Qwen 3-235B | 235 | Reasoning, MoE, open-source | April 2025 | Alibaba Cloud | 32 |
| Qwen 2.5-0.5B | 0.5 | IT, open-source | September 2024 | Alibaba Cloud | 32 |
| Qwen 2.5-3B | 3 | IT, open-source | September 2024 | Alibaba Cloud | 32 |
| Qwen 2.5-7B | 7 | IT, open-source | September 2024 | Alibaba Cloud | 131 |
| Qwen 2.5-14B | 14 | IT, open-source | September 2024 | Alibaba Cloud | 131 |
| Qwen 2.5-70B | 70 | IT, open-source | September 2024 | Alibaba Cloud | 131 |

**Supplementary Table S3:** Per-question inter-model decision entropy under zero-shot and agentic inference. This table reports Shannon entropy values, for the first 10 radiology questions of each dataset, under zero-shot and agentic retrieval-augmented inference. Entropy (H) was computed from the distribution of answer choices across the 34-model panel, with higher values indicating greater dispersion of model decisions and lower values indicating stronger concentration of answers. For each question, entropy under zero-shot and agentic inference is provided, along with the paired entropy change (ΔH), defined as agentic minus zero-shot entropy. Negative ΔH values indicate reduced dispersion and increased inter-model stability under agentic inference, whereas positive ΔH values indicate increased dispersion. Questions are grouped by dataset.

| Dataset | question ID | H Zero-shot | H Agentic | ΔH |
|---|---|---|---|
| **Benchmark-RadQA dataset** | | | |
| Benchmark-RadQA|1 | 1.08 | 0.94 | -0.14 |
| Benchmark-RadQA|2 | 1.05 | 0.13 | -0.92 |
| Benchmark-RadQA|3 | 0.13 | 0.00 | -0.13 |
| Benchmark-RadQA|4 | 0.00 | 0.13 | 0.13 |
| Benchmark-RadQA|5 | 1.02 | 0.90 | -0.11 |
| Benchmark-RadQA|6 | 0.93 | 0.95 | 0.02 |
| Benchmark-RadQA|7 | 0.63 | 0.13 | -0.50 |
| Benchmark-RadQA|8 | 0.71 | 0.00 | -0.71 |
| Benchmark-RadQA|9 | 0.49 | 0.35 | -0.14 |
| Benchmark-RadQA|10 | 0.52 | 0.22 | -0.29 |
| **Board-RadQA dataset** | | | |
| Benchmark-RadQA|12 | 0.00 | 0.00 | 0.00 |
| Benchmark-RadQA|13 | 0.72 | 0.13 | -0.58 |
| Benchmark-RadQA|14 | 0.40 | 0.30 | -0.10 |
| Benchmark-RadQA|15 | 0.26 | 0.00 | -0.26 |
| Benchmark-RadQA|16 | 0.00 | 0.00 | 0.00 |
| Benchmark-RadQA|17 | 0.00 | 0.00 | 0.00 |
| Benchmark-RadQA|18 | 0.00 | 0.00 | 0.00 |
| Benchmark-RadQA|19 | 0.35 | 0.13 | -0.22 |
| Benchmark-RadQA|20 | 1.13 | 0.35 | -0.77 |
| Benchmark-RadQA|21 | 0.35 | 0.00 | -0.35 |

**Supplementary Table S4:** Per-question majority agreement strength, correctness, and paired agreement changes under zero-shot and agentic inference. This table reports, for the first 10 radiology questions of each dataset, the majority fraction (M) under zero-shot and agentic retrieval-augmented inference, where M is defined as the proportion of the 34 models selecting the modal answer option. The paired change in agreement strength (ΔM) is calculated as agentic minus zero-shot majority fraction. Positive ΔM values indicate increased inter-model agreement under agentic inference, whereas negative values indicate decreased agreement. For each question, majority correctness (C) is additionally reported for both inference strategies, coded as 1 if the majority answer matches the reference standard and 0 otherwise. Questions are categorized according to whether agreement increased with a correct majority, increased with an incorrect majority, decreased, or remained unchanged. The absolute rank of non-zero ΔM values (used in the paired Wilcoxon signed-rank test) is provided to document the contribution of each question to the rank-based inference. Questions are grouped by dataset.

| Dataset \| question ID | M | | ΔM | C | | Category | Absolute rank of \|ΔM\| |
|---|---|---|---|---|---|---|---|
| | Zero-shot | Agentic | | Zero-shot | Agentic | | |
| **Benchmark-RadQA dataset** | | | | | | | |
| Benchmark-RadQA\|1 | 0.56 | 0.65 | 0.09 | 0 | 0 | Agreement ↑ & incorrect | 100.5 |
| Benchmark-RadQA\|2 | 0.62 | 0.97 | 0.35 | 1 | 1 | Agreement ↑ & correct | 166.5 |
| Benchmark-RadQA\|3 | 0.97 | 1.00 | 0.03 | 1 | 1 | Agreement ↑ & correct | 53.0 |
| Benchmark-RadQA\|4 | 1.00 | 0.97 | -0.03 | 1 | 1 | Agreement ↓ | 53.0 |
| Benchmark-RadQA\|5 | 0.50 | 0.65 | 0.15 | 1 | 0 | Agreement ↑ & incorrect | 126.0 |
| Benchmark-RadQA\|6 | 0.62 | 0.59 | -0.03 | 1 | 1 | Agreement ↓ | 53.0 |
| Benchmark-RadQA\|7 | 0.76 | 0.97 | 0.21 | 1 | 1 | Agreement ↑ & correct | 149.0 |
| Benchmark-RadQA\|8 | 0.79 | 1.00 | 0.21 | 1 | 1 | Agreement ↑ & correct | 149.0 |
| Benchmark-RadQA\|9 | 0.85 | 0.91 | 0.06 | 1 | 1 | Agreement ↑ & correct | 82.0 |
| Benchmark-RadQA\|10 | 0.85 | 0.94 | 0.09 | 1 | 1 | Agreement ↑ & correct | 100.5 |
| **Board-RadQA dataset** | | | | | | | |
| Board-RadQA\|1 | 1.00 | 1.00 | 0.00 | 1 | 1 | No change | 0.0 |
| Board-RadQA\|2 | 0.79 | 0.97 | 0.18 | 1 | 1 | Agreement ↑ & correct | 140.0 |
| Board-RadQA\|3 | 0.91 | 0.91 | 0.00 | 1 | 1 | No change | 0.0 |
| Board-RadQA\|4 | 0.94 | 1.00 | 0.06 | 1 | 1 | Agreement ↑ & correct | 82.0 |
| Board-RadQA\|5 | 1.00 | 1.00 | 0.00 | 1 | 1 | No change | 0.0 |
| Board-RadQA\|6 | 1.00 | 1.00 | 0.00 | 1 | 1 | No change | 0.0 |
| Board-RadQA\|7 | 1.00 | 1.00 | 0.00 | 1 | 1 | No change | 0.0 |
| Board-RadQA\|8 | 0.91 | 0.97 | 0.06 | 1 | 1 | Agreement ↑ & correct | 82.0 |
| Board-RadQA\|9 | 0.12 | 0.06 | -0.06 | 0 | 0 | Agreement ↓ | 72.5 |
| Board-RadQA\|10 | 0.91 | 1.00 | 0.09 | 1 | 1 | Agreement ↑ & correct | 100.5 |

**Supplementary Table S5:** Per-question robustness of correctness under zero-shot and agentic inference and robustness transition patterns. This table reports, for the first 10 radiology questions of each dataset, robustness of correctness under zero-shot and agentic retrieval-augmented inference. Robustness is defined as the fraction of the 34 models selecting the ground-truth answer. For each inference method, the total number of correct model outputs (out of 34), the corresponding robustness score (proportion correct), and the assigned robustness bin (low, moderate, high) are provided. The paired change in robustness (ΔR) is computed as agentic minus zero-shot robustness score. The transition column summarizes categorical shifts between robustness bins (for example, low → high, high → low, moderate → high, or no change), enabling identification of upward transitions, stability, and robustness collapse cases. Questions are grouped by dataset.

| Dataset \| question ID | Zero-shot | | | Agentic | | | Delta robustness | Transition |
|---|---|---|---|---|---|---|---|---|
| | Total correct (out of 34) [n] | Robustness score | Robustness bin | Total correct (out of 34) [n] | Robustness score | Robustness bin | | |
| **Benchmark-RadQA dataset** | | | | | | | | |
| Benchmark-RadQA \| 1 | 0 | 0.00 | Low | 7 | 0.21 | Low | 0.21 | No Change |
| Benchmark-RadQA \| 2 | 20 | 0.59 | Medium | 33 | 0.97 | High | 0.38 | Improved |
| Benchmark-RadQA \| 3 | 33 | 0.97 | High | 33 | 0.97 | High | 0.00 | No Change |
| Benchmark-RadQA \| 4 | 34 | 1.00 | High | 33 | 0.97 | High | -0.03 | No Change |
| Benchmark-RadQA \| 5 | 15 | 0.44 | Medium | 6 | 0.18 | Low | -0.26 | Decreased |
| Benchmark-RadQA \| 6 | 16 | 0.47 | Medium | 11 | 0.32 | Low | -0.15 | Decreased |
| Benchmark-RadQA \| 7 | 22 | 0.65 | Medium | 32 | 0.94 | High | 0.29 | Improved |
| Benchmark-RadQA \| 8 | 26 | 0.76 | Medium | 34 | 1.00 | High | 0.24 | Improved |
| Benchmark-RadQA \| 9 | 29 | 0.85 | High | 31 | 0.91 | High | 0.06 | No Change |
| Benchmark-RadQA \| 10 | 29 | 0.85 | High | 32 | 0.94 | High | 0.09 | No Change |
| **Board-RadQA dataset** | | | | | | | | |
| Board-RadQA \| 1 | 31 | 0.91 | High | 34 | 1.00 | High | 0.09 | No Change |
| Board-RadQA \| 2 | 22 | 0.65 | Medium | 33 | 0.97 | High | 0.32 | Improved |
| Board-RadQA \| 3 | 31 | 0.91 | High | 31 | 0.91 | High | 0.00 | No Change |
| Board-RadQA \| 4 | 31 | 0.91 | High | 34 | 1.00 | High | 0.09 | No Change |
| Board-RadQA \| 5 | 34 | 1.00 | High | 34 | 1.00 | High | 0.00 | No Change |
| Board-RadQA \| 6 | 34 | 1.00 | High | 34 | 1.00 | High | 0.00 | No Change |
| Board-RadQA \| 7 | 34 | 1.00 | High | 34 | 1.00 | High | 0.00 | No Change |
| Board-RadQA \| 8 | 24 | 0.71 | Medium | 33 | 0.97 | High | 0.26 | Improved |
| Board-RadQA \| 9 | 19 | 0.56 | Medium | 31 | 0.91 | High | 0.35 | Improved |
| Board-RadQA \| 10 | 31 | 0.91 | High | 34 | 1.00 | High | 0.09 | No Change |

**Supplementary Table S6:** Distribution of per-question robustness of correctness across a 34-model panel for the Benchmark-RadQA and Board-RadQA datasets. Robustness is defined as the fraction of models answering a question correctly. Questions are categorized into low (R < 0.4), medium (0.4 ≤ R < 0.8), and high (R ≥ 0.8) robustness. Values represent the percentage of questions in each category or transition type. Mean Δ robustness denotes the average paired change in robustness between agentic and zero-shot inference. P-values derive from paired Wilcoxon signed-rank tests on continuous robustness scores.

| Metric | | Pooled dataset | Benchmark-RadQA dataset | Board-RadQA dataset |
|---|---|---|---|---|
| Number of questions | | 169 | 104 | 65 |
| High robustness | Zero-shot | 50% | 47% | 54% |
| | Agentic | 72% | 63% | 85% |
| Medium robustness | Zero-shot | 41% | 39% | 44% |
| | Agentic | 17% | 21% | 10% |
| Low robustness | Zero-shot | 9% | 14% | 2% |
| | Agentic | 11% | 16% | 5% |
| Robustness increased | | 27% | 23% | 32% |
| Robustness decreased | | 7% | 9% | 5% |
| Robustness unchanged | | 66% | 68% | 63% |
| Mean Δ robustness | | 0.07 | 0.06 | 0.08 |
| Median Δ robustness | | 0.06 | 0.03 | 0.06 |
| P-value | | $5.6 \times 10^{-9}$ | $1.2 \times 10^{-4}$ | $7.4 \times 10^{-6}$ |
| Effect size (rank-biserial r) | | 0.45 | 0.37 | 0.56 |

**Supplementary Table S7:** Association between output verbosity and correctness across inference methods and datasets. This table summarizes response length (token count) distributions for correct and incorrect model outputs under zero-shot and agentic retrieval-augmented inference, stratified by dataset. Verbosity is reported as the median number of tokens with interquartile range (IQR) for correct and incorrect responses separately, along with the median difference (correct minus incorrect). All comparisons were performed within inference method to isolate verbosity-correctness relationships from systematic length differences induced by the inference pipeline. Statistical significance was assessed using two-sided Mann-Whitney U tests comparing correct vs. incorrect responses within each condition, with Cliff's δ reported as a non-parametric effect size. Counts reflect the total number of individual model responses analyzed per condition.

| Dataset | Pooled dataset | | Benchmark-RadQA dataset | | Board-RadQA dataset | |
|---|---|---|---|---|---|---|
| | Zero-shot | Agentic | Zero-shot | Agentic | Zero-shot | Agentic |
| Correct [n] | 4269 | 4664 | 2505 | 2713 | 1764 | 1951 |
| Incorrect [n] | 1477 | 1082 | 1031 | 823 | 446 | 259 |
| Median verbosity (Correct) | 280 | 660 | 282 | 702 | 279 | 596 |
| Median verbosity (Incorrect) | 256 | 668 | 259 | 689 | 250.5 | 593 |
| IQR (Correct) | 334 | 404 | 338 | 405 | 324 | 382 |
| IQR (Incorrect) | 295.0 | 375.8 | 318.0 | 353.0 | 265.0 | 441.5 |
| Median difference | 24.0 | 8.0 | 23.0 | 13.0 | 28.5 | 3.0 |
| P-value | 0.020 | 0.833 | 0.489 | 0.129 | 0.001 | 0.392 |
| Cliff's δ | 0.04 | -0.004 | 0.02 | 0.04 | 0.11 | 0.03 |

**Supplementary Table S8:** High-consensus, low-robustness failure cases. This table enumerates questions exhibiting coordinated but incorrect convergence under zero-shot or agentic inference. Cases were defined a priori as questions satisfying both a high inter-model consensus strength and low robust correctness, specifically a majority fraction $M_{q,m} \geq 0.8$ and a robustness score $R_{q,m} < 0.4$, where the majority fraction denotes the proportion of models selecting the modal answer option and the robustness score denotes the fraction of models selecting the reference-standard correct answer. These cases represent scenarios in which a large majority of models converged on the same answer despite most models being incorrect. Values are reported separately for zero-shot and agentic inference conditions.

| Dataset \| question ID | Method | Majority fraction (M) | Robustness (R) | Majority option | Correct answer |
|---|---|---|---|---|---|
| Benchmark-RadQA\|59 | agentic | 0.82 | 0.00 | A | D |
| Benchmark-RadQA\|60 | agentic | 0.82 | 0.15 | D | C |
| Benchmark-RadQA\|65 | zero-shot | 0.94 | 0.03 | A | C |
| Benchmark-RadQA\|65 | agentic | 0.82 | 0.03 | A | C |

**Supplementary Table S9:** Per-question clinical severity distribution and inter-rater agreement for incorrect answer options. This table reports, for the first 10 radiology questions of each dataset, the distribution of radiologist-assigned clinical severity labels for incorrect answer options and the corresponding inter-rater agreement statistics. For every question, all incorrect answer options selected by at least one model were evaluated independently by three clinicians and categorized as low, moderate, or high severity according to predefined clinical impact criteria. For each question the number of model outputs that were incorrect, the counts of incorrect model outputs linked to answer options assigned to each severity category after aggregation, the fraction of incorrect outputs within that question falling into each severity class are given. Inter-rater agreement across the three annotators is summarized per question using the observed agreement, expected agreement under chance, and Fleiss' κ. Fleiss' κ values close to zero indicate minimal agreement beyond chance. Questions are grouped by dataset (Benchmark-RadQA and Board-RadQA).

| Dataset \| question ID | Incorrect model outputs [n] | Observed agreement (P) | Expected agreement ($P_e$) | Fleiss' κ | Low severity [n] | Low severity Probability | Moderate severity [n] | Moderate severity Probability | High severity [n] | High severity Probability |
|---|---|---|---|---|---|---|---|---|---|---|
| **Benchmark-RadQA dataset** | | | | | | | | | | |
| Benchmark-RadQA\|1 | 3 | 0.33 | 0.67 | -1 | 1 | 0.33 | 2 | 0.67 | 0 | 0.00 |
| Benchmark-RadQA\|2 | 3 | 0.33 | 0.67 | -1 | 0 | 0.00 | 2 | 0.67 | 1 | 0.33 |
| Benchmark-RadQA\|3 | 3 | 0.33 | 0.67 | -1 | 0 | 0.00 | 2 | 0.67 | 1 | 0.33 |
| Benchmark-RadQA\|4 | 3 | 0.33 | 0.67 | -1 | 0 | 0.00 | 2 | 0.67 | 1 | 0.33 |
| Benchmark-RadQA\|5 | 3 | 0.33 | 0.67 | -1 | 0 | 0.00 | 3 | 1.00 | 0 | 0.00 |
| Benchmark-RadQA\|6 | 3 | 1.00 | 1.00 | 0 | 1 | 0.33 | 2 | 0.67 | 0 | 0.00 |
| Benchmark-RadQA\|7 | 3 | 0.33 | 0.67 | -1 | 0 | 0.00 | 2 | 0.67 | 1 | 0.33 |
| Benchmark-RadQA\|8 | 3 | 0.33 | 0.67 | -1 | 2 | 0.67 | 0 | 0.00 | 1 | 0.33 |
| Benchmark-RadQA\|9 | 3 | 0.33 | 0.67 | -1 | 2 | 0.67 | 1 | 0.33 | 0 | 0.00 |
| Benchmark-RadQA\|10 | 3 | 0.33 | 0.67 | -1 | 0 | 0.00 | 3 | 1.00 | 0 | 0.00 |
| **Benchmark-RadQA dataset** | | | | | | | | | | |
| Board-RadQA\|1 | 4 | 0.33 | 0.67 | -1 | 0 | 0.00 | 3 | 0.75 | 1 | 0.25 |
| Board-RadQA\|2 | 4 | 0.00 | 0.33 | -0.5 | 0 | 0.00 | 3 | 0.75 | 1 | 0.25 |
| Board-RadQA\|3 | 4 | 0.33 | 0.67 | -1 | 0 | 0.00 | 4 | 1.00 | 0 | 0.00 |
| Board-RadQA\|4 | 4 | 0.33 | 0.67 | -1 | 1 | 0.25 | 3 | 0.75 | 0 | 0.00 |
| Board-RadQA\|5 | 4 | 1.00 | 1.00 | 0 | 0 | 0.00 | 2 | 0.50 | 2 | 0.50 |
| Board-RadQA\|6 | 4 | 1.00 | 1.00 | 0 | 0 | 0.00 | 2 | 0.50 | 2 | 0.50 |
| Board-RadQA\|7 | 4 | 0.33 | 0.67 | -1 | 0 | 0.00 | 1 | 0.25 | 3 | 0.75 |
| Board-RadQA\|8 | 4 | 1.00 | 1.00 | 0 | 4 | 1.00 | 0 | 0.00 | 0 | 0.00 |
| Board-RadQA\|9 | 4 | 0.33 | 0.67 | -1 | 1 | 0.25 | 3 | 0.75 | 0 | 0.00 |
| Board-RadQA\|10 | 4 | 0,33 | 0,67 | -1 | 4 | 1,00 | 0 | 0,00 | 0 | 0,00 |

# Supplementary Note 1

*Dataset-based results*

## Agentic reasoning alters inter-model decision stability

To assess whether stability shifts were consistent across datasets, we repeated the entropy analysis within Board-RadQA (n = 65) and Benchmark-RadQA (n = 104) (**Table 2, Figure 2**). In Board-RadQA, entropy was lower under agentic inference, with the median decreasing from 0.35 (IQR 0.50) to 0.13 (IQR 0.13) and the mean from 0.41 to 0.20. The paired shift was significant (P = $9.6 \times 10^{-6}$); rank-biserial r = −0.94), with median ΔH = −0.13 and mean ΔH = −0.20. In Benchmark-RadQA, the median decreased from 0.53 (IQR 0.56) to 0.24 (IQR 0.57) and the mean from 0.55 to 0.38. The paired shift was also significant (P = $9.4 \times 10^{-9}$); r = −0.64), with median ΔH = −0.13 and mean ΔH = −0.18. The direction of change was predominantly stabilizing in both datasets, but heterogeneity remained. In Board-RadQA, entropy decreased for 46/65 questions (71%), increased for 8/65 (12%), and was unchanged for 11/65 (17%). In Benchmark-RadQA, entropy decreased for 69/104 (66%), increased for 23/104 (22%), and was unchanged for 12/104 (12%).

**Supplementary Table S3** illustrates the range of per-question behavior in concrete terms. For example, Benchmark-RadQA includes several large stabilizations (e.g., question 2: H 1.05 → 0.13, ΔH = −0.92; question 8: H 0.71 → 0.00, ΔH = −0.71), but also occasional destabilization (e.g., question 4: H 0.00 → 0.13, ΔH = +0.13) and near-null change (e.g., question 6: H 0.93 → 0.95, ΔH = +0.02). Board-RadQA shows multiple near-ceiling agreement cases with H = 0 under both methods (e.g., questions 1 and 5–7), alongside substantial stabilizations when baseline dispersion exists (e.g., question 9: H 1.13 → 0.35, ΔH = −0.77; question 2: H 0.72 → 0.13, ΔH = −0.58). These examples underscore that agentic inference often compresses answer distributions, but does not impose a uniform direction of change on every question. As in the main text, these stability shifts are correctness-agnostic and should be interpreted as changes in coordination structure rather than validity.

## Dataset-stratified consensus strength and correctness

We evaluated majority-vote agreement strength separately within Board-RadQA (n = 65) and Benchmark-RadQA (n = 104) to assess consistency across datasets (**Table 3, Supplementary Table S4, Figure 3**). In both datasets, majority fractions were higher under agentic inference and paired shifts were significant when restricting the Wilcoxon signed-rank test to non-zero ΔM pairs. Board-RadQA showed a median majority fraction increase from 0.91 (IQR 0.18) under zero-shot to 0.97 (IQR 0.06) under agentic inference, with median ΔM = 0.03 (P = $1.5 \times 10^{-4}$). Benchmark-RadQA increased from 0.82 (IQR 0.27) to 0.94 (IQR 0.21), with a larger median ΔM = 0.06 (P = $5.7 \times 10^{-7}$). These results indicate that agreement strengthening under agentic inference was present in both datasets, with a larger typical gain in Benchmark-RadQA.

The distribution of agreement-shift categories further highlights dataset differences in how consensus changes relate to correctness. In Board-RadQA, agreement increased with a correct majority in 40/65 questions (62%), increased with an incorrect majority in 1/65 (2%), decreased in 10/65 (15%), and showed no change in 14/65 (22%). In Benchmark-RadQA, agreement increased with a correct majority in 55/104 (53%), increased with an incorrect majority in 10/104 (10%), decreased in 20/104 (19%), and was unchanged in 19/104 (18%). Thus, while both datasets showed more agreement increases than decreases, Benchmark-RadQA had a higher proportion of cases where agreement increased around an incorrect majority.

**Supplementary Table S4** provides per-question examples that illustrate these patterns. For instance, Benchmark-RadQA includes cases where agreement increases substantially with a correct majority (e.g., question 2: M 0.62 → 0.97, ΔM = +0.35) and cases where agreement increases while the majority remains incorrect (e.g., question 1: M 0.56 → 0.65, ΔM = +0.09, majority incorrect under both methods; question 5: M 0.50 → 0.65, ΔM = +0.15, majority correctness shifts from correct to incorrect). Board-RadQA contains many ceiling or near-ceiling cases with no change (e.g., question 1: M = 1.00 in both conditions) as well as typical agreement increases with correct majorities (e.g., question 2: M 0.79 → 0.97, ΔM = +0.18). Together, these stratified results complement the pooled finding: agentic inference generally strengthens consensus, but the correctness implications of stronger agreement remain question- and dataset-dependent, and agreement amplification can also occur around incorrect decisions.

# Agentic reasoning increases robustness of correctness across models

Dataset-stratified analyses revealed consistent but heterogeneous robustness shifts (**Supplementary Tables S5** and **S6**). In the Benchmark-RadQA dataset (104 questions), mean robustness increased from 0.71 to 0.77 and the median from 0.79 to 0.94. The share of high-robustness questions rose from 47% to 63%, primarily driven by medium-to-high transitions, while low-robustness questions remained non-trivial (13% to 15%). The paired robustness shift was significant (P = $1.2 \times 10^{-4}$; r = 0.37). In the Board-RadQA dataset (65 questions), baseline robustness was higher. Mean robustness increased from 0.80 to 0.88 and the median from 0.85 to 0.97. High-robustness questions increased from 54% to 85%, with a marked reduction in the medium category (45% to 11%), while low-robustness cases remained rare but increased slightly in absolute terms. The paired effect was significant (P = $7.4 \times 10^{-6}$; r = 0.56).

Thus, both datasets showed upward robustness shifts, but the magnitude and transition structure differed.

# Robustness gains coexist with rare but severe collapse cases

Dataset-level breakdown clarifies the structure of these collapse events (**Supplementary Tables S5** and **S6**). In the Benchmark-RadQA dataset, 9 of 104 questions (9%) showed reduced robustness. These included several medium-to-low or high-to-medium transitions with ΔR between −0.06 and −0.50. In the Board-RadQA dataset, 3 of 65 questions (5%) exhibited robustness decreases, including one medium-to-low collapse with ΔR = −0.74. These events reflect coordinated model shifts under shared structured retrieval. Although infrequent, their magnitude underscores that robustness gains coexist with structured tail-risk behavior.

## Output verbosity is a weak and inconsistent proxy for correctness

In the Benchmark-RadQA dataset, verbosity was not associated with correctness under either inference strategy. Under zero-shot inference, median verbosity was 282 tokens for correct responses and 259 for incorrect responses (P = 0.489; Cliff's δ = 0.02). Under agentic inference, median verbosity was 702 tokens for correct responses and 689 tokens for incorrect responses (P = 0.129; Cliff's δ = 0.04). Effect sizes were negligible in both cases. In the Board-RadQA dataset, zero-shot inference showed a statistically significant but small separation, with median verbosity of 279 tokens for correct responses and 250.5 tokens for incorrect responses (P = 0.001; Cliff's δ = 0.11). This difference did not persist under agentic reasoning (596 vs. 593 tokens; P = 0.392; Cliff's δ = 0.03). Thus, verbosity-correctness associations were weak, dataset-dependent, and not strengthened by agentic retrieval.

## Consensus strength and robust correctness are only partially coupled

Dataset-stratified analyses revealed differences in coupling strength. Under zero-shot inference, correlations were strong in both datasets (Benchmark-RadQA: ρ = 0.93, P = $1.4 \times 10^{-45}$; Board-RadQA: ρ = 0.81, P = $2.3 \times 10^{-16}$). Under agentic reasoning, the association strengthened further in Benchmark-RadQA (ρ = 0.96, P = $4.0 \times 10^{-56}$) but weakened in Board-RadQA (ρ = 0.69, P = $1.6 \times 10^{-10}$). Large separations between correct and incorrect majorities were observed in both datasets (all Mann-Whitney comparisons significant; Cliff's δ large). Median majority agreement was consistently higher for correct than for incorrect questions in both datasets and inference modes. These findings indicate that while pooled coupling remains strong, agentic reasoning modifies the strength of consensus-robustness alignment in a dataset-dependent manner.

# Supplementary Note 2

*Examples of high-consensus/low-robustness failures*

To clarify what high-consensus but low-robustness cases look like qualitatively, we provide two representative examples from Benchmark-RadQA.

In question 65, the stem stated that endobronchial biopsy of a hypermetabolic lung lesion showed adenocarcinoma of gastrointestinal origin (CDX-2 positive), while also describing a second lesion elsewhere on imaging. Most large models converged on "chest wall metastasis of rectal adenocarcinoma," explicitly citing the biopsy and immunohistochemistry as decisive evidence. The benchmark label, however, was "encapsulated fat necrosis." The majority answer was incorrect because models implicitly assumed the question referred to the biopsied thoracic lesion, collapsing the differential around the biopsy finding and excluding benign options. The convergence therefore reflects prompt-induced framing bias driven by a highly salient diagnostic detail, rather than a shared incorrect medical fact or retrieval artifact.

In question 60, the stem described a patient with osteosarcoma who became hypotensive immediately after major orthopedic surgery, and CT angiography showed multiple bilateral filling defects in the pulmonary arteries. Most models selected "thromboembolic pulmonary embolism," citing postoperative status, malignancy-associated thrombosis risk, hemodynamic instability, and the nonspecific appearance of arterial filling defects. The benchmark label was "pulmonary tumor embolism." Because the imaging description did not include discriminating features between thrombotic and tumor emboli, models defaulted to the more common peri-operative diagnosis. Here, convergence reflects a shared base-rate heuristic under ambiguous imaging evidence rather than shared misinformation.

These cases demonstrate that coordinated incorrect convergence can arise when multiple models rely on the same salient cue or default reasoning strategy under structural ambiguity, even when overall consensus strength is high.

# Supplementary Note 3

## *Datasets and question formatting*

A total of 169 text-only multiple-choice radiology questions were evaluated across two expert-curated datasets.

The Benchmark-RadQA dataset contains 104 questions and is derived from the RadioRAG study through combination of RSNA-RadioQA and ExtendedQA. RSNA-RadioQA contributes 80 questions curated from peer-reviewed cases in the Radiological Society of North America Case Collection. Questions were authored from clinical history and imaging-description text contained in the original cases, and images were intentionally excluded to enforce a text-only setting. Differential diagnoses explicitly listed by original case authors were removed during curation to reduce direct answer leakage. The source dataset spans 18 radiology subspecialties.

ExtendedQA contributes 24 questions that were written by a radiologist and validated by a second board-certified radiologist to probe generalization and reduce contamination risk. Because ExtendedQA was originally open-ended, it was converted to multiple-choice while preserving the original stem and reference-standard answer. To harmonize scoring across Benchmark-RadQA, each question was standardized to four options by adding three clinically plausible distractors. Distractor candidates were generated using OpenAI GPT-4o and o3 as draft generators, after which review and finalization were performed by a board-certified radiologist to ensure plausibility, appropriate difficulty, and the absence of misleading or implausible options. These draft-generation models were used only to propose distractors and were not involved in determining ground truth[2].

The Board-RadQA dataset contains 65 board-style questions developed and validated by board-certified radiologists to reflect domains aligned with German radiology board certification. Questions were derived from representative cases and core concepts emphasized in the Technical University of Munich radiology curriculum, with the aim of minimizing overlap with online case collections and known training corpora. Board-RadQA questions were formatted as five-option multiple-choice items. Each item across both datasets has a single reference-standard correct option; Benchmark-RadQA[2] uses four options (A–D) and Board-RadQA uses five options (A–E).

# Supplementary Note 4

## *Agentic retrieval-augmented reasoning: knowledge base, tools, and orchestration workflow*

Under the agentic inference condition, each evaluated model is provided with a structured evidence report generated by a fixed retrieval-and-synthesis pipeline. Final answer selection is always produced by the evaluated model rather than by the orchestration engine, allowing isolation of how models use identical retrieved evidence.

Retrieval within agentic pipeline was restricted to Radiopaedia.org, a peer-reviewed and openly accessible radiology knowledge base. Radiopaedia was selected as the exclusive retrieval domain to prioritize clinically validated content and avoid heterogeneous quality across open web sources. Domain restriction was enforced by appending site-restricted filters to all queries and by performing explicit post-retrieval domain verification. Retrieval was executed through a locally hosted SearXNG meta-search engine deployed in a containerized Docker environment to support consistency and reproducibility. Retrieved results were deduplicated and formatted into standardized evidence bundles for synthesis. The orchestration workflow was adapted from LangChain's Open Deep Research pipeline and implemented as a stateful directed graph using LangGraph. A supervisor component coordinated the workflow and delegated option-specific evidence gathering to research components. For each question, a neutral research plan with one section per answer option was constructed, option-specific tasks were assigned, and completed sections were synthesized into a final report containing an objective introduction and conclusion.

Each research module conducts targeted retrieval for a single option. Retrieval begins with focused queries using core option terms, followed by contextual queries incorporating salient clinical details from the question stem. When retrieval is insufficient, query refinement through simplification or synonym substitution is performed. Retrieval attempts are capped at four per option, and failure to obtain sufficient evidence is documented. Each option section summarizes retrieved evidence and explicitly notes supporting and contradicting points with traceable citations to Radiopaedia sources.

A diagnostic abstraction step guides retrieval. For each question, the Mistral Large model generates a concise comma-separated summary of key clinical concepts. This summary is used only internally for query formulation and is not shown to evaluated models. The keyword-extraction template was:

"Extract and summarize the key clinical details from the following radiology question. Provide a concise, comma-separated summary of keywords and key phrases in one sentence only. Question: {question_text}.

Summary:"

The orchestration engine responsible for planning and report writing was implemented using OpenAI GPT-4o-mini for tool use and synthesis. This orchestration model does not select the final answer. To reduce benchmark exposure during report generation, the orchestration model was provided only derived keywords and the answer options (not the verbatim stem) for query formulation; evaluated models still receive the full stem and options during evaluation. Because identical reports are provided to all evaluated models, no preferential advantage is conferred to any model family.

# Supplementary Note 5

## *Inference prompts, scoring, and validation procedures*

Each model answered each question under two conditions: zero-shot and agentic retrieval-augmented reasoning. In the zero-shot condition, only the question and options were provided. In the agentic condition, the same question and options were preceded by the structured Radiopaedia-derived report.

Standardized prompt templates were used.

Zero-shot template:

"You are a highly knowledgeable medical expert… {question} {options}"

Agentic template:

"You are a highly knowledgeable medical expert… {report} {question} {options}"

Responses were scored against the reference-standard option. Automated option matching was applied for scalability. When multiple options were listed, correctness was assigned only when the correct option was explicitly and unambiguously selected. Manual spot checks confirmed agreement between automated scoring and human interpretation.

Constrained verification was additionally performed using Mistral Large as a binary adjudicator. The model received the response and reference answer and returned "Yes" or "No" indicating presence of the correct answer. To validate this procedure, 20 questions across five representative models (DeepSeek-R1, o3, Llama3.3-70B, MedGemma-27B-text-it, Qwen 2.5-3B) were manually reviewed, totaling 100 responses. Complete concordance between manual and automated verification was observed.